\documentclass{article}

\usepackage{microtype}
\usepackage{graphicx}
\usepackage{subcaption}
\usepackage{booktabs} %
\usepackage{makecell}
\usepackage{hyperref}
\usepackage{caption}

\usepackage{xspace}

\newcommand{\RETURN}{\STATE \textbf{return} }
\usepackage[preprint]{icml2026}

\definecolor{cvprblue}{rgb}{0.21,0.49,0.74}

\hypersetup{
    pagebackref,
    breaklinks,  
    colorlinks=true,
    linkcolor=cvprblue,
    citecolor=cvprblue,
    urlcolor=cvprblue
}

\usepackage{fontawesome5} %
\usepackage{amsmath}
\usepackage{amssymb}
\usepackage{mathtools}
\usepackage{amsthm}

\newcommand{\opensource}{\textcolor{teal}{\faGithub}\hspace{0.3em}} 

\newcommand{\closedsource}{\textcolor{red!75!black}{\faLock}\hspace{0.3em}}

\newcommand{\agent}{\textcolor{blue}{\faRobot}\hspace{0.3em}}
\usepackage{multirow}
\usepackage{widetext}
\usepackage{titletoc}
\usepackage[capitalize,noabbrev]{cleveref}
\usepackage[most]{tcolorbox}
\usepackage{pifont}
\definecolor{agent-orange}{RGB}{184, 115, 51} 
\definecolor{question-gray}{RGB}{60, 60, 60}
\usepackage{wrapfig}
\definecolor{OA_O}{HTML}{FF8888} %
\definecolor{OA_m}{HTML}{FFB075} %
\definecolor{OA_n1}{HTML}{FFCC44} %
\definecolor{OA_i}{HTML}{DBC520} %
\definecolor{OA_A}{HTML}{A5D555} %
\definecolor{OA_g}{HTML}{70E095} %
\definecolor{OA_e}{HTML}{55D5C5} %
\definecolor{OA_n2}{HTML}{4AC0D5} %
\definecolor{OA_t}{HTML}{44AADD} %
\usepackage{twemojis}
\newcommand{\OmniAgent}{%
\textcolor{OA_O}{O}%
\textcolor{OA_m}{m}%
\textcolor{OA_n1}{n}%
\textcolor{OA_i}{i}%
\textcolor{OA_A}{A}%
\textcolor{OA_g}{g}%
\textcolor{OA_e}{e}%
\textcolor{OA_n2}{n}%
\textcolor{OA_t}{t}%
}
\theoremstyle{plain}

\theoremstyle{definition}

\theoremstyle{remark}

\definecolor{MacaronGold}{HTML}{F6D04D}
\usepackage[textsize=tiny]{todonotes}
\usepackage[table]{xcolor}
\definecolor{lyellow}{HTML}{F7F2CC}
\definecolor{lpink}{HTML}{F6D6D6}
\definecolor{mycol1}{HTML}{F2EEEC}
\definecolor{mycol2}{HTML}{7E6A60}
\usepackage[most]{tcolorbox}
\usepackage{markdown}
\definecolor{promptorange}{RGB}{255, 153, 51} 

\newtcolorbox{promptbox}{
    colback=mycol2!10!white,           %
    colframe=mycol2!70!white,     %
    arc=8pt,                 %
    boxrule=0.8pt,           %
    left=10pt,               %
    right=10pt,              %
    top=10pt,                %
    bottom=10pt,             %
    upperbox=visible,
}
\newcommand{\var}[1]{\textcolor{promptorange}{\{#1\}}}

\newtcolorbox{mytocbox}[1][]{
    colback=mycol2!10!white,           %
    colframe=mycol2!70!white,  %
  title={Appendix Content},          %
  arc=2mm,                 %
  boxrule=1pt,             %
  #1
}

\icmltitlerunning{OmniAgent: Active Perception Agent for Omnimodal Audio-Video Understanding}
\begin{document}
\twocolumn[
\icmltitle{Active Perception Agent for Omnimodal Audio-Video Understanding}
\icmlsetsymbol{equal}{*}

\begin{icmlauthorlist}
\icmlauthor{Keda Tao}{zju,wl,ant}
\icmlauthor{Wenjie Du}{wl}
\icmlauthor{Bohan Yu}{ant}
\icmlauthor{Weiqiang Wang}{ant}
\icmlauthor{Jian liu}{ant}
\icmlauthor{Huan Wang}{wl}
\end{icmlauthorlist}

\centering
\icmlaffiliation{zju}{Zhejiang University}
\icmlaffiliation{wl}{Westlake University}
\icmlaffiliation{ant}{Ant Group}

\icmlcorrespondingauthor{Huan Wang}{wanghuan@westlake.edu.cn}
\icmlcorrespondingauthor{Jian Liu}{rex.lj@antgroup.com}

\icmlkeywords{Machine Learning, ICML}

\vskip 0.3in
]

\begin{widetext}
\begin{center}
\centering
\captionsetup{font={small}, skip=-1pt}
\vspace{-9mm}
\captionsetup{skip=3pt}
\includegraphics[width =1\linewidth]{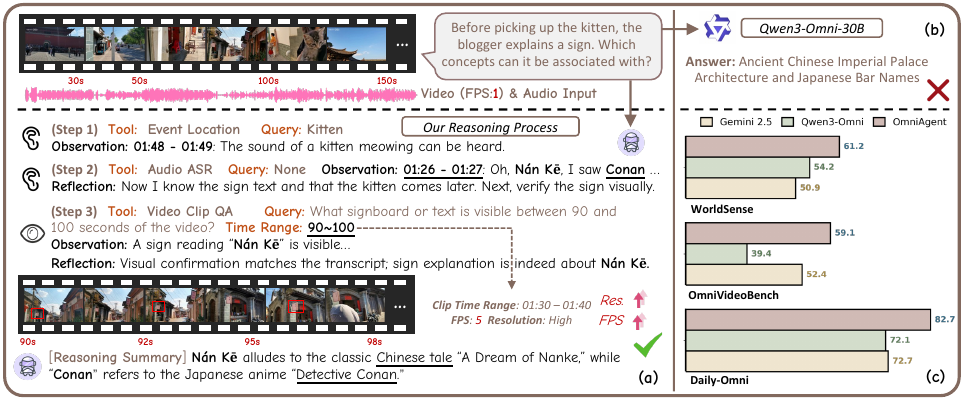}

\captionof{figure}{\textbf{(a)}: Illustration of \OmniAgent, an active perception agent designed for omnimodal understanding. Given a user query, our agent employs a recursive \emph{``Think-Act-Observe-Reflect''} loop and actively orchestrates multimodal tools (video, audio, and event tools) for fine-grained audio-video understanding.
Due to the unavailability of the thinking process, we use reflection in the final summary of the agent to better understand.
The presented video clip is from a vlog; the question is about two Chinese characters on a hanging signboard in the video. 
Initially, the agent utilizes audio to locate the temporal segment with the key information (``the kitten''), then invokes the video clip tool within that time window. 
Within the salient segment, we can afford model inference at an \emph{increased} spatial and temporal resolution.
With sufficient relevant visual evidence and the audio as input, the agent derives the correct answer.
\textbf{(b)}: In contrast, the end-to-end model Qwen3-Omni~\cite{xu2025qwen3} cannot achieve such fine-grained reasoning and gives the wrong answer.
\textbf{(c)}: Performance comparison on three audio-video understanding benchmarks. OmniAgent demonstrates superior performance without training, consistently outperforming strong end-to-end OmniLLMs such as Qwen3-Omni and Gemini 2.5-Flash~\cite{comanici2025gemini}.}
\label{fig:teaser}
\end{center}

\end{widetext}

\printAffiliationsAndNotice{}  %

\begin{abstract}
Omnimodal large language models have made significant strides in unifying audio and visual modalities; however, they often face challenges in fine-grained cross-modal understanding and have difficulty with multimodal alignment. 
To address these limitations, we introduce \textbf{OmniAgent}, to our best knowledge, the first fully active perception agent that dynamically orchestrates specialized unimodal tools to achieve more fine-grained omnimodal reasoning.  
Unlike previous works that rely on rigid, static workflows and dense frame-captioning, we demonstrate a paradigm shift from passive response generation to active multimodal inquiry. OmniAgent employs dynamic planning to autonomously orchestrate tool invocation on demand, strategically concentrating perceptual attention on task-relevant cues.
Central to our approach is a novel coarse-to-fine audio-guided perception paradigm, which leverages audio cues to localize temporal events and guide subsequent reasoning.
Extensive empirical evaluations on three audio-video understanding benchmarks demonstrate that OmniAgent achieves state-of-the-art performance, surpassing leading open-source and closed-source models by substantial margins of 10\% - 20\% accuracy without training.
\end{abstract}

\section{Introduction}

\begin{figure}[t!]
 \centering
\includegraphics[width=1\linewidth]{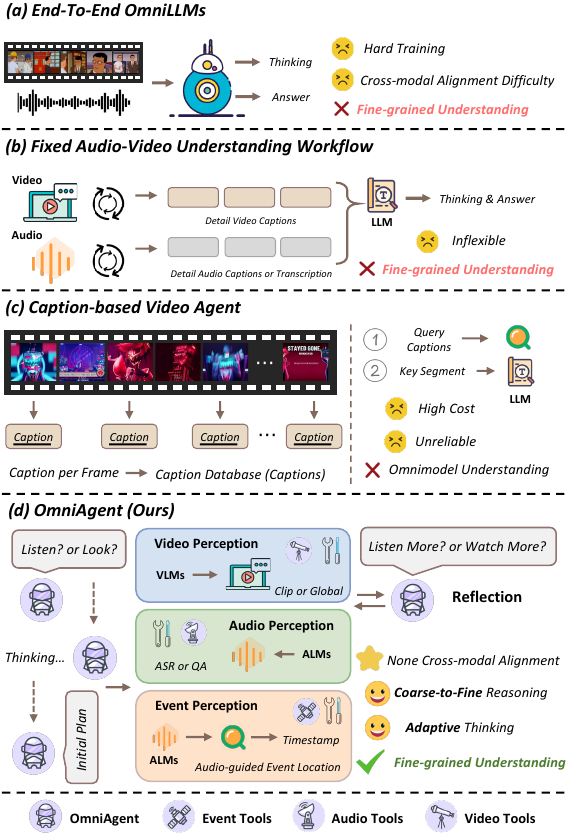}
\caption{
\emph{(a) End-to-end OmniLLMs} implicitly fuse modalities but suffer from high training costs, difficult alignment, and limited fine-grained reasoning. %
\emph{(b) Fixed workflow agents} rely on rigid pipelines, lacking the flexibility to allocate attention for fine-grained analysis adaptively. 
\emph{(c) Caption-based agents} incur high precomputation costs and noise sensitivity, often failing to capture comprehensive multimodal context.
\emph{(d) Our OmniAgent} employs active perception reasoning and inquiry. Within an iterative, reflective loop, the agent strategically leverages the ability to understand video and audio.
This explicitly solves the cross-modal alignment difficulty and achieves fine-grained understanding.
}
\label{fig:motivation}
\vspace{-7mm}
\end{figure}

Recently, end-to-end omnimodal large language models (OmniLLMs) have achieved encouraging results by integrating visual and audio encoders into a unified architecture~\cite{tang2025video,zhang2024videoin,xu2025qwen2,xu2025qwen3,yang2025humanomniv2,ge2025arc,shu2025audio,yang2025audio,shu2025audio,ai2025ming}. 
Despite this progress, as shown in \cref{fig:motivation}(a), OmniLLMs still face challenges in fine-grained cross-modal understanding, and the joint alignment training of audio and video representations poses significant challenges~\cite{galougah2025aura,chowdhury2024meerkat,sun2023fine}.
Consequently, models often cannot respond accurately.

A critical empirical observation is that while MLLMs have demonstrated exceptional proficiency in unimodal tasks~\cite{wang2025internvl3,team2025kimi,bai2025qwen2,wang2024qwen2,chu2023qwen,ding2025kimi,bai2025qwen3vl,feng2025can,feng2025rewardmap}, cross-modal understanding remains constrained by the challenges of temporal and feature alignment~\cite{chowdhury2024meerkat,jiang2025specific,fan2025fine}. 
Consequently, developing an agent to synergize the capabilities of distinct modalities is now a promising direction. 
Previous omnimodal agents rely predominantly on static workflows~\cite{cao2025xgc,zhou2025daily}, as illustrated in \cref{fig:motivation}(b).
These methods face challenges in effectively harnessing the inherent reasoning capabilities of models for dynamic planning, thereby impeding the attainment of a fine-grained understanding.

Recent works have yielded significant advancements in agent-based video-only understanding~\cite{zhang2025deep,zhang2024omagent,fan2024videoagent,wang2024videoagent,yang2025vca,wang2025videotree,pang2025mr,wang2025active,chowdhury2025magnet,yin2025videoarm,tian2025ego}.
Specifically, temporal event localization is paramount for fine-grained analysis. 
Prevailing approaches predominantly rely on frame-captioning, where captions are generated for sampled frames, stored, and subsequently retrieved and analyzed iteratively by agents~\cite{wang2024videoagent,zhang2025deep,wang2025videotree}. 
While these methods refine their hypotheses through multi-step inference, they incur substantial computational overhead. Moreover, the generated captions may occasionally be irrelevant to the query.
However, in the context of audio-visual understanding, the audio modality presents distinct challenges yet offers a unique opportunity: unlike redundant visual signals, audio naturally provides accurate and concise temporal grounding information regarding the salient events~\cite{guo2025aligned,tao2025omnizip,wu2025survey,chen2025chronusomni,xie2025caption}. 
This information is efficiently utilized for event localization and to emulate a reasoning process akin to human cognition, thereby facilitating a more comprehensive cross-modal understanding.

Building on these insights, we present \OmniAgent, an agent specifically designed for omnimodal (audio-visual) understanding in an active reasoning fashion, which treats strong single-modal models as callable tools.
In contrast to previous methods constrained by fixed workflows, as shown in \cref{fig:motivation}(d), our approach initiates a fundamental paradigm shift from \emph{passive} response generation to \emph{active} information inquiry. 
The agent employs an LLM as the central component to orchestrate tool invocation, determining the optimal modality to use -- it explicitly decides \emph{whether} to attend to audio or video, and \emph{how} to process the information. Note, this process is completely \emph{autonomous}, decided by the LLM itself. The tool calling and decision making is transparent to us, thus explainable and optimizable.
By strategically selecting spatial and temporal focus range (\emph{i.e.}, deciding precisely where to look and listen), OmniAgent achieves genuine, fine-grained cross-modal understanding.

Specifically, we construct a comprehensive library of tools categorized into three distinct sets: (1) Video tools, (2) Audio tools, and (3) Event tools. 
The \emph{video toolset} enables global captioning and general visual QA, while also allowing for the analysis of specific temporal windows at higher sampling rates to support more fine-grained understanding. 
The \emph{audio toolset} incorporates audio captioning and detailed QA capabilities, complemented by timestamped automatic speech recognition (ASR) for precise speech grounding. 
Within the \emph{event toolset}, we propose a novel \emph{audio-based event localization} strategy. 
This mechanism empowers the agent to autonomously query and temporally localize events across the entire audio stream, establishing temporal anchors for subsequent fine-grained analysis.
By synthesizing the capabilities of distinct MLLMs, our agent adaptively leverages their complementary strengths to facilitate joint, fine-grained analysis, utilizing cross-modal corroboration to maximize audio-visual comprehension performance.

Extensive experiments show that OmniAgent achieves the \textit{best} accuracy on several audio-video understanding benchmarks, surpassing the state-of-the-art open-source and closed-source models, such as Qwen3-Omni-30B~\cite{xu2025qwen3} and Gemini2.5-Flash~\cite{comanici2025gemini}, by a significant margin of 10-20\% accuracy without training.

The main contributions of this work are:
\begin{itemize}
    \item We introduce \emph{OmniAgent}, a novel agent-based framework tailored for comprehensive audio-video understanding. Employing an active perception strategy, it dynamically modulates attention between auditory and visual modalities, and, via a self-reflective mechanism, solves the cross-modal alignment problem.
    \item We construct a comprehensive modality-specific toolkit and introduce an audio-guided event localization algorithm designed to facilitate \emph{fine-grained cross-modal reasoning}.
    \item Experimental results on several audio-video understanding benchmarks show that OmniAgent achieves the new SoTA, with significant accuracy improvement compared with open-source and closed-source models.
\end{itemize}

\section{Related Work}

\subsection{Omnimodal Large Language Models}
End-to-end OmniLLMs aim to achieve a common understanding across all modalities, including image, audio, video, and text.
By leveraging multimodal data, these architectures acquire richer contextual representations and gain a deeper understanding of inter-modal relationships~\cite{xu2025qwen2,tong2025interactiveomni,xu2025qwen3,xie2024mini,tang2025video,zhang2024videoin,yang2025humanomniv2,ge2025arc,shu2025audio,sun2024video,li2024baichuanomni,team2025longcat}.
Recent works, such as Qwen3-Omni~\cite{xu2025qwen2} and the Video-SALMONN series~\cite{sun2024video,tang2025video}, have introduced state-of-the-art end-to-end models capable of unified multimodal perception. 
Among closed-source models, Gemini~\cite{comanici2025gemini,team2024gemini,team2023gemini} stands as a powerful baseline, distinguished by its strong multimodal understanding capabilities. 
However, end-to-end models face significant hurdles: they require complex alignment training across multiple modalities and often face challenges to achieve fine-grained cross-modal understanding.

\begin{figure*}
    \centering
    \includegraphics[width=1\linewidth]{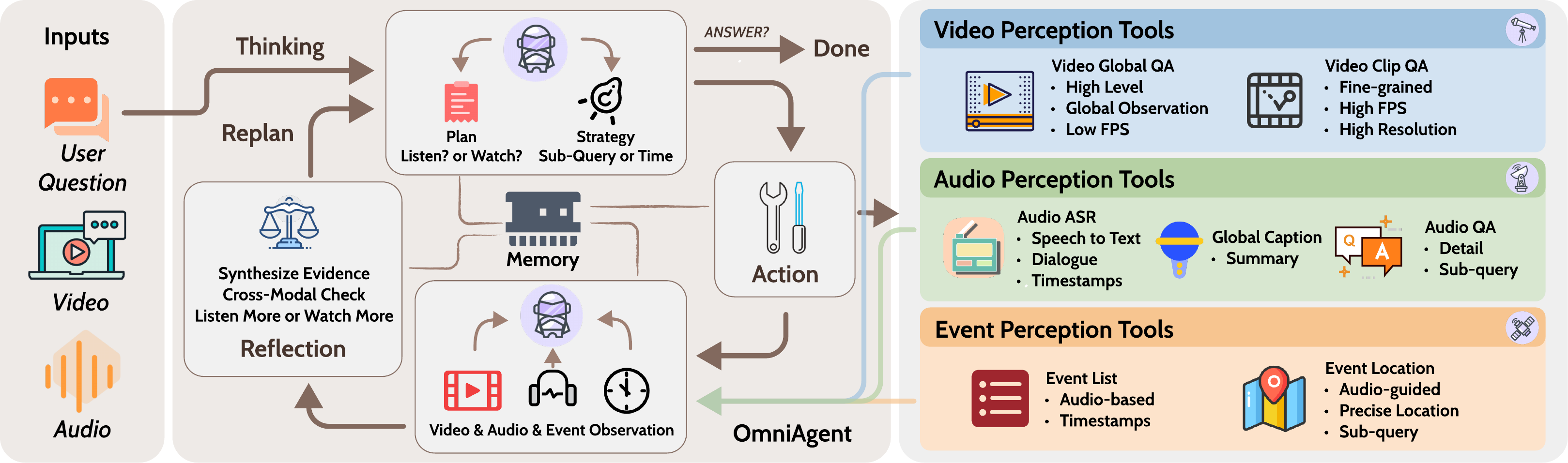}
    \caption{\textbf{Overview of the OmniAgent framework.} The system processes audio and video inputs through an iterative \emph{thinking-action-observe-reflection} cycle. The agent utilizes a comprehensive suite of perception tools (video, audio, and event) to gather fine-grained evidence, while the reflection module synthesizes observations to update the memory and decide whether to rethink or conclude the task.}
    \label{fig:method}
    \vspace{-4mm}
\end{figure*}

\subsection{Video Understanding Agent}
Leveraging the advanced capabilities of MLLMs, recent studies have investigated agentic approaches to address the intricacies of video understanding with video clip captioning~\cite{wang2024videoagent,zhang2024simple,jeoung2024adaptive,wang2025videochat,park2024too,ma2025drvideo,kahatapitiya2025language,jeoung2024adaptive,kugo2025videomultiagents}.
Concurrently, other methodologies have focused on decomposing complex queries into multi-step processes utilizing specialized tool modules~\cite{fan2024videoagent,liu2025videomind,min2024morevqa,zhu2025active,zhang2024omagent}. 
More recently, research has shifted away from static workflows to explore active agentic perception~\cite{yao2022react,yuan2025videodeepresearch,zhang2025deep,wang2025active,gao2025agentic,yang2025streamagent}, thereby enhancing long-form video comprehension. 
However, comprehensive audio-video understanding remains challenging due to the complexities of cross-modal alignment and fine-grained reasoning. 
Addressing this gap, we introduce OmniAgent to this holistic multimodal context for the first time. 
Departing from the rigid workflows or frame-captioning strategy, we propose a novel active mechanism that utilizes an audio-guided reasoning process.

\section{OmniAgent}

We introduce the \emph{OmniAgent}, specifically designed for omnimodal audio-video understanding. 
In contrast to conventional paradigms that rely on passive frame processing or rigid execution protocols, 
OmniAgent functions as an active perception. It dynamically orchestrates a suite of modality-specific perception tools, effectively reformulating audio-video understanding from a passive retrieval task into an active, sequential decision-making process.
This approach circumvents the alignment bottlenecks inherent in end-to-end models and achieves fine-grained understanding. 
To the best of our knowledge, OmniAgent is the \emph{first} active perception agent framework for omnimodal understanding.

\subsection{Overview and Problem Formulation}

\textbf{Motivation.}
Existing end-to-end OmniLLMs typically process video and audio streams by projecting them into a shared latent space. 
However, this paradigm exhibits a fundamental limitation: the inability to allocate attentional resources between modalities adaptively. 
Crucially, query-relevant information is often modality-specific; valid responses may hinge exclusively on auditory cues or demand scrutiny of high-resolution visual details. 
Constrained by fixed token budgets and joint optimization objectives, OmniLLMs lack the architectural flexibility to dynamically prioritize specific modalities or adjust processing granularity.
This deficiency frequently results in the degradation of fine-grained understanding.

\textbf{Formulation.} 
To address this, we formulate omnimodal understanding not as a static task, but as a sequential, active decision-making process. 
Let $\mathcal{V}$ and $\mathcal{A}$ denote the visual and audio streams, and $q$ be the user query. 
We define an agent $\pi$ and store a memory $\mathcal{M} = \{a_0, o_0, \dots, a_T, o_T\}$.
At each step $t$, the agent assesses its state $s_t$ and actively selects an action $a_t \in \mathcal{T}$ and gets an observation $o_t$ (e.g., \textit{Listen} to a segment or \textit{Watch} a specific region), to maximize the information gain regarding $q$. 
By explicitly decoupling the modalities into callable tools, OmniAgent empowers the model to autonomously determine the optimal modality and granularity—deciding when to rely on low-cost auditory cues and when to demand high-cost visual inspection—thereby solving the cross-modality alignment difficulty.

\subsection{Modality-Aware Expert Toolset}

To facilitate precise interaction with the environment, we devise a comprehensive toolset, denoted as $\mathcal{T}$, stratified by both modality and granularity. 
Functioning as the perceptual interfaces of the agent, these tools offer varying degrees of information density and computational overhead.

\textbf{\textcolor{mycol2}{\faWrench} Video Perception Tools ($\mathcal{T}_{\text{V}}$).} 
While visual processing yields rich semantic information, it incurs high computational costs. 
Relying solely on global representations often leads to the loss of fine-grained granular details. 
To address these trade-offs, we design two distinct visual tools: \textbf{\texttt{Global\_QA}}: $\mathcal{T}_{\text{VGA}}$ and \textbf{\texttt{Clip\_QA}}: $\mathcal{T}_{\text{VCA}}$. 
For $\mathcal{T}_{\text{VGA}}$, we employ sparse frame sampling to mitigate the overhead of long sequences.
Additionally, this tool allows the agent to identify initial visual cues for coarse temporal localization.
Conversely, $\mathcal{T}_{\text{VCA}}$ serves as the fine-grained analysis engine. 
It extracts video slices within a target temporal window and employs a higher sampling rate and input resolution. 
This enables deep visual reasoning, facilitating the detailed analysis of object actions and spatial relationships, while maintaining a balanced computational budget.

\textbf{\textcolor{mycol2}{\faWrench} Audio Perception Tools ($\mathcal{T}_{\text{A}}$).} 
Audio signals provide dense, complementary information essential for holistic video reasoning. 
First, the \textbf{\texttt{ASR}}: $\mathcal{T}_{\text{ASR}}$ transcribes spoken dialogue into text with precise timestamp alignment. 
This capability is indispensable for queries dependent on specific verbal cues or semantic narratives conveyed through speech. 
Second, the \textbf{\texttt{Global\_Caption}}: $\mathcal{T}_{\text{AGC}}$ synthesizes a summary of the acoustic environment, establishing a global auditory context.
Finally, the \textbf{\texttt{Audio\_QA}}: $\mathcal{T}_{\text{AQ}}$ tool empowers the agent to formulate targeted inquiries, extracting specific acoustic details required for understanding.

\begin{figure}[t]
    \centering
    \includegraphics[width=1\linewidth]{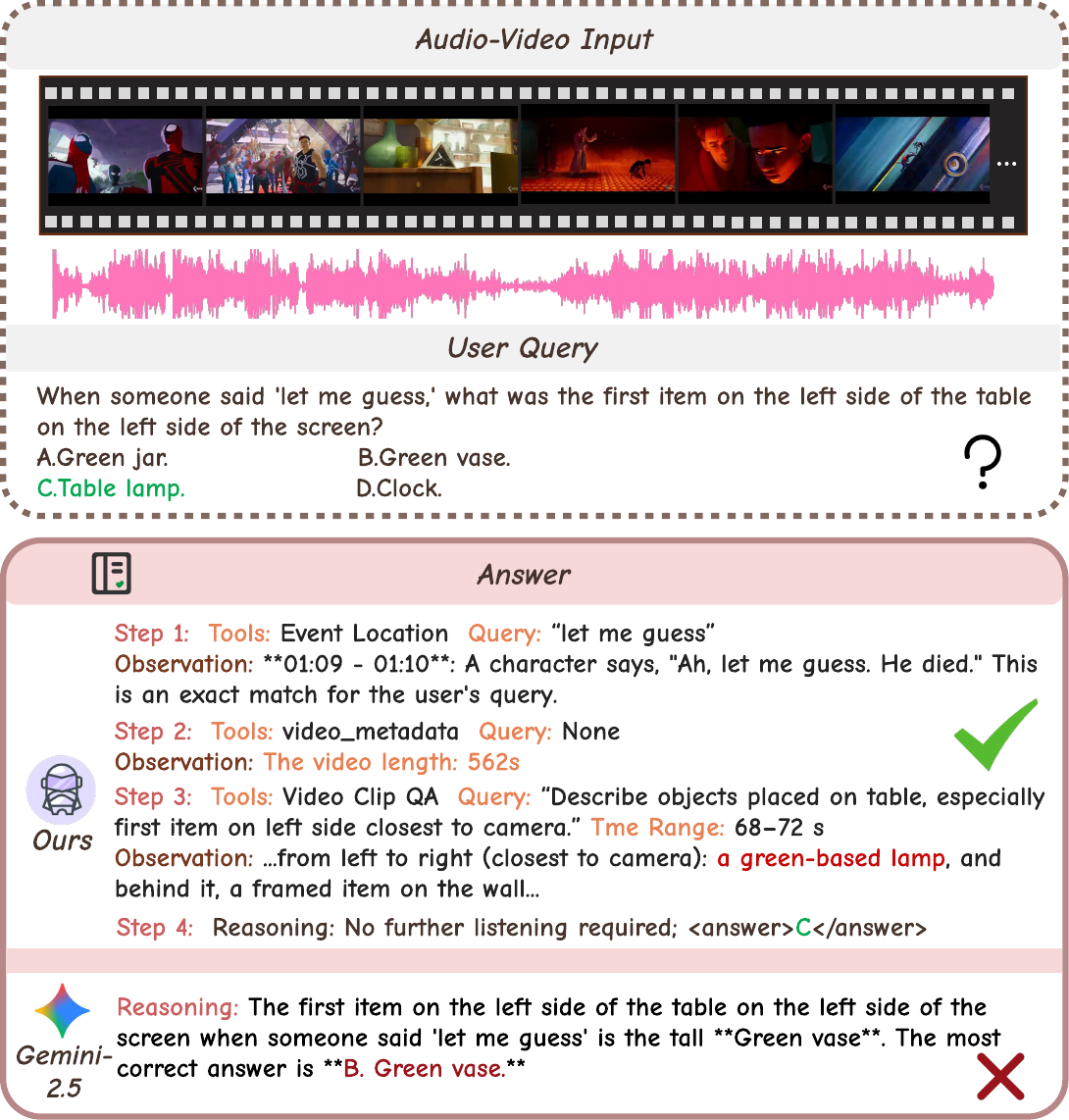}
    \vspace{-5mm}
    \caption{Visualization of the responses and underlying reasoning processes generated by our OmniAgent and Gemini2.5-Flash to an audio-video understanding question.}
    \label{fig:tool_event}
    \vspace{-5.5mm}
\end{figure}

\textbf{\textcolor{mycol2}{\faWrench} Event Perception Tools ($\mathcal{T}_{\text{E}}$).}
Fine-grained video understanding faces significant hurdles due to the computational prohibitiveness of high-frame-rate sampling over long sequences, rendering precise event localization a persistent challenge.
We identify the audio modality as a pivotal opportunity to address this bottleneck; unlike video, audio captures global context and event semantics with low cost. 
Leveraging this efficiency, we propose an audio-guided event localization method. 
Specifically, the \textbf{\texttt{Event List}}: $\mathcal{T}_{\text{EL}}$ tool processes the entire audio stream to extract a discrete list of detectable sound events, enabling the agent to discern the global semantic context. 
Complementarily, \textbf{\texttt{Event Location}}: $\mathcal{T}_{\text{ELO}}$ accepts specific queries to return precise timestamps.
This serves as an effective temporal proposal mechanism, allowing the agent to pinpoint occurrence times efficiently. 

\subsection{Agentic Design}
To fully exploit the intrinsic reasoning and planning capabilities of the LLMs, we eschew rigid workflows or prescriptive tool usage. Instead, we formulate an iterative \emph{Think-Act-Observe-Reflect} cycle, empowering the agent to actively orchestrate reasoning, planning, and execution across both modalities (\emph{audio and video}). As shown in \cref{fig:method}.

\textbf{Active Thinking.} 
Upon receiving an input query, our agent formulates a strategic inference plan designed to maximize accuracy while utilizing information from both modes. 
Crucially, the system assesses the cross-modal dependency of the query to prioritize the optimal modality dynamically—determining whether to employ a ``listen" or ``watch" strategy—and selects the appropriate retrieval tools accordingly. 
In the step $t$, we have:
\begin{equation}
a_t,\mathrm{args}_t=\pi_{\mathrm{plan}}(q,\mathcal{M}_t),
\end{equation}
where the agent maintains a comprehensive contextual memory $\mathcal{M}_t$, aggregating the initial query, sequential observations, and accumulated background information to facilitate robust multi-step reasoning. $\mathrm{args}_t$ is the corresponding action parameter of $a_t$, such as the sub-question for $\mathcal{T}_{\text{AQ}}$ or $\mathcal{T}_{\text{VGA}}$ and the video clip start time \& end time for $\mathcal{T}_{\text{VCA}}$.

\begin{algorithm}[t]
\caption{OmniAgent Inference Process}
\textbf{Input:} User Query $q$, Audio $\mathcal{A}$, Video $\mathcal{V}$, Toolset $\mathcal{T}$ \\
\textbf{Output:} Answer $y$
\begin{algorithmic}[1]
\STATE Initialize Memory $\mathcal{M}_0 \leftarrow \emptyset$
\WHILE{not \textit{Answered}}
    \STATE $a_t, \text{args}_t \leftarrow \pi_{\mathrm{plan}}(q,\mathcal{M}_t)$
    
    \IF{$a_t$ is \texttt{ANSWER}}
        \RETURN $\text{args}_t$
    \ENDIF
    
    \STATE $o_t \leftarrow \text{ExecuteTool}(a_t, \text{args}_t, \mathcal{A}, \mathcal{V})$
        
    \STATE $\text{Reflect}(q, \mathcal{M}_t, o_t)$
    \STATE $\mathcal{M}_{t+1} \leftarrow \mathcal{M}_t \cup \{ (a_t, o_t) \}$
    \STATE $t \leftarrow t + 1$
\ENDWHILE
\end{algorithmic}
\label{alo:agent}
\end{algorithm}

\begin{table*}[t]
\centering
\caption{\textbf{Comparison of different models on the Daily-Omni benchmark}. The \textbf{best} result among token pruning methods for each metric is in bold, and the \underline{second-best} is underlined. `A' denotes the incorporation of audio modalities, whereas `V' indicates reliance on visual modalities extracted from video inputs. The symbol `$*$' signifies that the model employs Chain-of-Thought (CoT) reasoning or extended inference before generating an answer.
}
\vspace{-2mm}
\resizebox{\textwidth}{!}{%
\begin{tabular}{lc|ccccccccc} 
\toprule
\makecell{Method} & 
\makecell{Modality} & 
\makecell{AV Event\\Alignment} & 
\makecell{Comparative} & 
\makecell{Context\\Understanding} & 
\makecell{Event\\Sequence} & 
\makecell{Inference} & 
\makecell{Reasoning} & 
\makecell{30s\\Subset} & 
\makecell{60s\\Subset} & 
\makecell{Avg} \\
\midrule
\multicolumn{11}{c}{\textbf{\emph{Closed-source Models}}} \\
\midrule

GPT-4o  & V & 47.90 & 62.60 & 52.33 & 52.61 & 66.23 & 66.29 & 55.64 & 57.45 & 56.47 \\
Qwen3-VL-Plus & V & 51.68 & 77.10 & 61.66 & 66.99 & 71.43 & 68.57 & 63.68 & 66.55 & 65.00\\
\rowcolor{mycol1}
Gemini 2.0-Flash    & A+V & 62.18 & 73.28 & 63.73 & 63.72 & 76.62 & 75.43 & 67.23 & 68.55 & 67.84 \\
\rowcolor{mycol1}
Gemini 2.5-Flash$^*$ & A+V & - & - & - & - & - & - & - & - & \underline{72.70} \\
\midrule
\multicolumn{11}{c}{\textbf{\emph{Open-source Models}}} \\
\midrule

Unified-IO-2 XXL-8B &A+V & 25.63 & 31.30 & 26.42 & 25.82 & 35.06 & 29.71 & 26.74 & 30.00 & 28.24 \\
VideoLLaMA2-7B & A+V & 35.71 & 35.88 & 35.75 & 31.70 & 40.91 & 34.29 & 38.02 & 31.82 & 35.17 \\
Qwen2.5-Omni-7B & A+V & 44.12 & 51.15 & 38.86 & 40.52 & 57.79 & 61.71 & 46.68 & 48.36 & 47.45 \\
Ola-7B  & A+V & 40.34 & 61.07 & 40.41 & 43.46 & 63.64 & 69.71 & 51.47 & 49.82 & 50.71 \\
\rowcolor{mycol1}
Qwen3-Omni-30B & A+V & 61.90 & \underline{79.25} & \underline{69.47} & \underline{65.32} & 82.67 & \underline{85.92} & 71.28 & \underline{74.29} & 72.08 \\
\midrule
\multicolumn{11}{c}{\textbf{\emph{Agent-based Methods}}} \\
\midrule
DVD & V & 49.32 & 57.47 & 57.12 & 58.45 & 70.37 & 63.24 & 56.31 & 62.41 & 59.22 \\
Daily-Omni & A+V & 51.68 & 68.70 & 60.10 & 53.92 & 78.57 & 71.43 & 63.99 & 59.27 & 61.82 \\
\rowcolor{mycol1}
XGC-AVis & A+V & \underline{63.50} & 77.10 & 68.40 & 64.40 & \textbf{85.10} & 82.30 & \underline{71.60} & 71.50 & 71.50 \\
\rowcolor{mycol2!20!white}
OmniAgent (Ours) & A+V & \textbf{80.67} & \textbf{83.21} & \textbf{80.83} & \textbf{81.05} &\underline{83.36} & \textbf{86.86} & \textbf{80.37} & \textbf{85.45} & \textbf{82.71} \\
\bottomrule
\end{tabular}%
}
\label{tab:main_dailyomni}

\vspace{-4mm}
\end{table*}

\textbf{Action \& Observation} 
The selected tool by the agent is executed on the corresponding modality streams:
\begin{equation}
    o_t=\mathrm{Execute}(a_t,\mathrm{args}_t,\mathcal{V},\mathcal{A}),
\end{equation}
where $o_t$ is the output of tools with text response or timestamp. 
The model will update the perception of the entire audio and video based on the initial thought derived from the observation of both modalities.
And the memory is updated: $\mathcal{M}_{t+1} = \mathcal{M}_t \cup \{(a_t, o_t)\}$.

\textbf{Reflection \& Rethinking.} 
Before the next iteration, the agent critically assesses all the acquired evidence. 
It determines the efficacy of the executed tool by synthesizing current outputs with historical context, using this data to refine the execution plan dynamically. 
Crucially, the module executes a cross-modal consistency check to identify potential discrepancies between visual and auditory signals.
Consequently, if the agent determines that the accumulated multimodal evidence is insufficient to resolve the query, or if a cross-modal discrepancy arises necessitating further exploration, it reinitiates the thinking cycle. 
This iterative loop persists until the \textbf{\texttt{ANSWER}} operation is explicitly invoked, at which point the system synthesizes a final response and summary addressing the original user query.

Notably, \emph{while omniagent predominantly leverages audio for event localization to achieve better time alignment, our framework can execute precise visual event localization when necessary and achieve better performance in unimodal-based reasoning} (see \cref{fig:case_study_only_video,fig:case_study_only_audio} in the Appendix~\ref{sec:case_study_uni} for more case analyses.). 
\cref{fig:tool_event} illustrates the reasoning process of our agent.
In summary, we equip the agent with an evidence-based, reflective, and flexible action execution mechanism. 
Mirroring human cognitive processes, the system selectively extracts multimodal information and reasons about modal perception sub-problems.
Consequently, this approach circumvents the computational complexity associated with rigid, dense cross-modal alignment. 
Through this iterative reasoning process, OmniAgent achieves fine-grained cross-modal understanding by synthesizing perceptions from both modalities, ultimately delivering superior accuracy in response to the given question.

\section{Experimental Results}
\label{sec:experimental_results}

\subsection{Experimental Settings} \label{sec:experimental_settings}
\noindent \textbf{Benchmarks.}
We evaluate our method on three widely-used audio-video understanding benchmarks: Daily-Omni~\cite{zhou2025daily}, OmniVideoBench~\cite{li2025omnivideobench}, and WorldSense~\cite{hong2025worldsense}. 
Daily-Omni primarily evaluates performance on short-form video segments with durations of 30s and 60s, whereas OmniVideoBench comprehensively assesses audio-visual understanding capabilities in long-form videos. Complementarily, WorldSense gauges multimodal comprehension across eight distinct domains, focusing specifically on medium-length videos.

\begin{table}[t!]
\caption{\textbf{Quantitative comparison on the OmniVideoBench}. The \textbf{best} result among token pruning methods for each metric is in bold, and the \underline{second-best} is underlined. `A' denotes the incorporation of audio modalities, whereas `V' indicates reliance on visual modalities extracted from video inputs. The symbol `$*$' signifies that the model employs reasoning or extended inference before generating an answer.
}
\setlength{\tabcolsep}{4pt}
\vspace{-2mm}
\resizebox{\linewidth}{!}{%
\begin{tabular}{lc|ccccc}
\toprule
\multicolumn{1}{c}{Method} & \multicolumn{1}{c}{Modality} & \multicolumn{1}{c}{(0,1{]} min} & \multicolumn{1}{c}{(1,5{]} min} & \multicolumn{1}{c}{(5,10{]} min} & \multicolumn{1}{c}{(10,30{]} min} & \multicolumn{1}{c}{Avg.} \\ 
\midrule
\multicolumn{7}{c}{\textbf{\emph{Closed-source Models}}}  \\
\midrule
Qwen3-VL-Plus & V & 36.92 & 45.27 & 37.87 & 30.65 & 38.93 \\
\rowcolor{mycol1}
Gemini-2.0-Flash & A+V & 49.40 & 43.15 & 41.05 & 34.87 & 41.50 \\
\rowcolor{mycol1}
Gemini-2.5-Flash$^*$ & A+V & \underline{55.42} & \underline{55.10} & \underline{47.37} & \underline{52.11} & \underline{52.40} \\
\midrule
\multicolumn{7}{c}{\textbf{\emph{Open-source Models}}}  \\
\midrule
Qwen2.5-VL-72B &  V & 33.13 & 30.03 & 31.88 & 24.43 & 29.50 \\
VideoLLaMA2-7B &  A+V & 32.00 & 28.20 & 29.60 & 28.29 & 29.20 \\
Qwen2.5-Omni-7B &  A+V & 41.57 & 27.41 & 25.33 & 26.72 & 29.30  \\
Baichuan-Omni-1.5 &  A+V & 28.92 & 31.78 & 28.38 & 32.44 & 30.70  \\
\rowcolor{mycol1}
Qwen3-Omni-30B  &  A+V & 45.78 & 37.03 & 38.86 & 35.11 & 38.40  \\
\midrule
\multicolumn{7}{c}{\textbf{\emph{Agent-based Methods}}}  \\
\midrule
\rowcolor{mycol2!20!white}
OmniAgent (Ours) & A+V & \textbf{66.08} & \textbf{58.53} & \textbf{59.03} & \textbf{55.64} & \textbf{59.10} \\
\bottomrule
\end{tabular}
}
\label{tab:omnivideobench}
\vspace{-6mm}
\end{table}

\noindent \textbf{Compared Methods.}
We compare our agent with  open-source MLLMs: VideoLLaMA2~\cite{damonlpsg2024videollama2}, Ola~\cite{liu2025ola}, Unified-IO-2~\cite{lu2024unified}, Qwen2.5-Omni~\cite{xu2025qwen2}, Qwen2.5-VL~\cite{bai2025qwen2}, Baichuan-Omni-1.5~\cite{li2025baichuan}, 
video-SALMONN 2 ~\cite{sun2024video,tang2025video}, Qwen3-VL~\cite{bai2025qwen3vl} and the state-of-the-art model Qwen3-Omni~\cite{xu2025qwen3}. And various closed-source MLLMs: Gemini2.5-Flash~\cite{comanici2025gemini}, GPT-4o~\cite{openai2023gpt4}, Gemini2.0-Flash, and OpenAI o3.
In addition, we also compare with the SoTA agent-based audio-video understanding framework: Daily-Omni~\cite{zhou2025daily} and Xgc-avis~\cite{cao2025xgc}. 
These approaches are predicated on static or semi-rigid agent workflows to varying degrees, standing in distinct contrast to the dynamic adaptability of our proposed framework.
We also compare with the video understanding agent DVD~\cite{zhang2025deep}.

\begin{table*}[t]
\centering
\captionsetup{skip=4pt}
\caption{\textbf{Comparison of different models on the WorldSense}. We compare our agent-based (\agent) method against various baselines, including closed-source (\closedsource) and open-source (\opensource) models. The \textbf{best} result among token pruning methods for each metric is in bold.}
\setlength{\tabcolsep}{10pt}
\resizebox{1\linewidth}{!}{
\begin{tabular}{lcc|ccccccccc} 
\toprule
Method & \#Params & Modality & \begin{tabular}[c]{@{}c@{}}Tech \&\\ Science\end{tabular} & \begin{tabular}[c]{@{}c@{}}Culture \&\\ Politics\end{tabular} & Daily Life & \begin{tabular}[c]{@{}c@{}}Film \&\\ TV\end{tabular} & Performance & Games & Sports & Music & Avg. \\ \midrule
\closedsource GPT-4o & - & V & 48.0 & 44.0 & 38.3 & 43.5 & 41.9 & 41.2 & 42.6 & 42.7 & 42.6 \\ 
\closedsource Gemini 1.5 Pro & - & A+V & 53.7 & 47.2 & 50.3 & 50.4 & 52.4 & 46.8 & 40.2 & 42.0 & 48.0 \\
\rowcolor{mycol1}
\closedsource Gemini 2.5 Flash & - & A+V & 55.1 & 48.2 & 53.0 & 48.8 & 56.2 & 47.2 & 46.3 & 50.0 & 50.9 \\ 
\midrule
\opensource Qwen2.5-Omni & 7B & A+V & 47.8 & 49.8 & 43.6 & 43.8 & 48.3 & 39.1 & 43.5 & 47.3 & 45.4 \\
\rowcolor{mycol1}
\opensource video-SALMONN 2+ & 72B & A+V & 59.0 & 63.1 & 54.0 & 59.9 & 58.1 & 54.1 & 51.9 & 54.4 & 56.5 \\
\rowcolor{mycol1}
\opensource Qwen3-Omni & 30B & A+V & - & - & - & - & - & - & - & - & 54.0 \\ 
\midrule
\rowcolor{mycol2!20!white}
\agent OmniAgent & - & A+V & \textbf{64.3} & \textbf{66.3} & \textbf{59.4} & \textbf{63.1} & \textbf{62.2} & \textbf{59.2} & \textbf{55.8} & \textbf{60.3} & \textbf{61.2} \\
\bottomrule
\end{tabular}
}

\label{tab:WorldSense}
\vspace{-1mm}
\end{table*}

\begin{figure*}[t!]
    \centering
    \includegraphics[width=\linewidth]{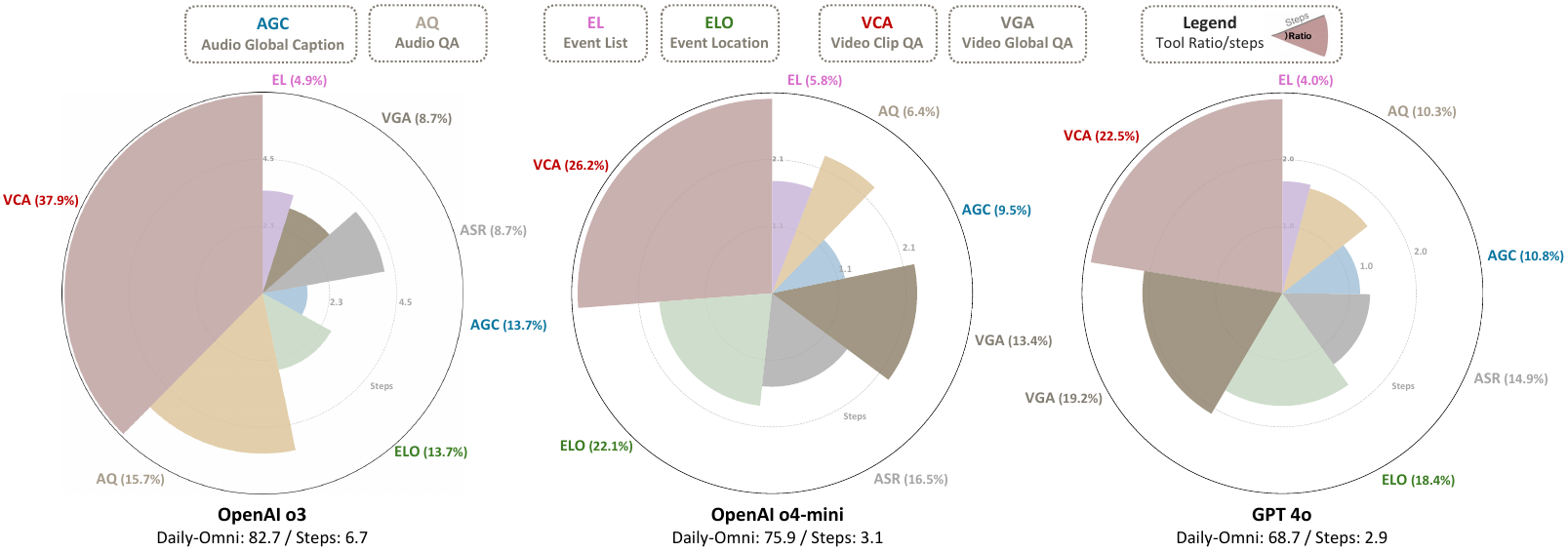}
    \caption{Analysis of the behavior of OmniAgent with different core LLM models. We quantified tool utilization patterns by calculating both the proportion of invocations (call ratio) and the average number of reasoning steps per call. In the resulting visualization, the sector angle represents the \emph{tool call ratio}, and the magnitude of the radius denotes the specific execution steps at which the tool was invoked.}
    \label{fig:analysis_tool_llm}
    \vspace{-3mm}
\end{figure*}

\textbf{Implementation Details.}
\label{sec:implementation_details}
For the core of the agent, we use OpenAI o3 as the brain because of its excellent reasoning capabilities.
We restrict the maximum number of iteration steps to 30. 
Regarding the component modules, we employ Qwen3-VL as the backbone for video perception and utilize Qwen3-Omni for audio global caption and ASR. 
Additionally, we select Gemini-2.5-Flash for the event perception tool and $\mathcal{T}_{\text{AQ}}$, capitalizing on its better time grounding capabilities. 
For more information, see Appendix~\ref{sec:appendix_implementation}.

\subsection{Comparison with SoTA Models}

\cref{tab:main_dailyomni} presents the results on the Daily-Omni Benchmark.
Specifically, OmniAgent substantially outperforms both proprietary baselines, such as Gemini-2.5-Flash-Thinking (72.7\%), and state-of-the-art open-source models like Qwen3-Omni (72.08\%), achieving a remarkable overall accuracy of \emph{82.71\%}. 
This result validates that our agentic framework, by effectively synergizing specialized uni-modal capabilities, circumvents the inherent challenges of rigid cross-modal alignment. 
Furthermore, it demonstrates the efficacy of audio guidance in enhancing fine-grained audio-visual understanding. 
Relative to competing agent-based architectures, OmniAgent yields performance gains of \emph{10\%-20\%}, underscoring the critical value of its self-planning, self-reflection, and inquiry mechanisms.

For long video evaluation and for more difficult questions, \cref{tab:omnivideobench} presents the results on the OmniVideoBench. The enhancement of OmniAgent compared to Qwen3-Omni-30B is notably significant, achieving an overall accuracy rate of \emph{59.1\%}. This performance substantially surpasses that of other open-source and closed-source end-to-end models, thereby further validating the efficacy of our agent algorithm. 
\cref{fig:tool_event} illustrates the comparative inference capabilities of OmniAgent
 against Gemini2.5-Flash, demonstrating that OmniAgent effectively resolves complex queries by leveraging active cross-modal reasoning.
For medium-length video in the WorldSense benchmark, as shown in \cref{tab:WorldSense}, it can also reflect the leading position of our OmniAgent. In the Appendix~\ref{sec:case_study}, we provide more case studies about agent reasoning in different inputs.

\subsection{Analyses on Reasoning Behaviors}
\label{sec:main_analyses_llm}

The core LLM serves as the central reasoning engine within our OmniAgent. 
It autonomously synthesizes multimodal information to dynamically orchestrate tool execution. 
To elucidate these mechanisms, we conducted a quantitative analysis of tool invocation patterns across various LLMs on the Daily-Omni Benchmark~\cite{zhou2025daily}.
Specifically, we measured both the distribution of tool calls and the associated average reasoning steps, as illustrated in \cref{fig:analysis_tool_llm}.
In the Appendix~\cref{sec:appendix_behavior}, we provide more behavior analyses and show the behavior of the agent in different benchmarks.

\textbf{Finding 1.} Across all LLM backbones, we observe a consistent strategic pattern: agents prioritize $\mathcal{T}_{\text{AGC}}$ (AGC) in the \emph{initial phase} to establish a global contextual background. 
Subsequently, the reasoning process culminates with $\mathcal{T}_{\text{VCA}}$ (VCA), employed to extract the precise, fine-grained evidence necessary for the \emph{final response}. 
This distinct sequential progression—from global audio context to localized visual verification—empirically validates the efficacy of our algorithmic design and the utility of the provided toolset.

\textbf{Finding 2.}  
The behavior of using OpenAI o3 as the LLM in our agent design aligns precisely with the core objective of cross-modal fine-grained understanding. 
We observe that the model preferentially utilizes granular tools—specifically $\mathcal{T}_{\text{VCA}}$ and $\mathcal{T}_{\text{AQ}}$—during the final resolution phase, while strategically deploying $\mathcal{T}_{\text{ELO}}$ for event localization during the intermediate reasoning stages. 
Conversely, computationally intensive tools ($\mathcal{T}_{\text{ELO}}$ and $\mathcal{T}_{\text{VGA}}$) provide macro-level insights, as they are unable to provide granular, fine-grained details.
This progression effectively exemplifies the coarse-to-fine cognitive flow orchestrated by our agentic algorithm.

\textbf{Finding 3.} 
OpenAI o4-mini and GPT-4o exhibit a propensity for rapid convergence, frequently bypassing deeper reflection and iterative inspection phases. 
This tendency is particularly pronounced in GPT-4o, which demonstrates an excessive reliance on coarse-grained $\mathcal{T}_{\text{VGA}}$ outputs. Consequently, it fails to interrogate fine-grained visual details, resulting in low accuracy.
Similarly, o4-mini displays a significant modality bias, disproportionately prioritizing visual information while neglecting the exploration of audio.

\textbf{Insight.} 
GPT-4o yields suboptimal performance, largely attributable to premature convergence on coarse-grained evidence. 
In contrast, o3 demonstrates a more deliberative process, effectively leveraging both modalities to unearth fine-grained details. 
This disparity underscores the critical necessity of the thinking-to-reflection cycle. 
Furthermore, our findings suggest that agentic systems must actively mitigate modal biases and strive for cross-modal consensus. 
Design protocols should prevent dominant visual signals from overshadowing critical auditory cues, thereby ensuring a holistic and accurate understanding of the video.

\begin{table}[t]
    \centering
    \caption{Ablation on model choices for signal modal toolsets. In this experiment, we use Qwen2.5-Omni-7B and Gemini2.5-Flash for evaluation.}
    \vspace{-1.5mm}
    \setlength{\tabcolsep}{9pt}
    \resizebox{\linewidth}{!}{
    \begin{tabular}{ccc|c}
        \toprule
        \multicolumn{3}{c|}{\textbf{Tools}} & \textbf{Daily-Omni} \\
        \textit{Video Tool} & \textit{Audio Tool} & \textit{Event Tool} & (Avg.) \\
        \midrule
        \rowcolor{mycol2!20!white}
        Qwen3-VL & Qwen3-Omni & Gemini-2.5 & 82.7 \\
        Qwen3-VL & Qwen3-Omni & Gemini-2.0 & 74.2 \\
        Qwen2.5-VL & Qwen3-Omni & Qwen3-Omni & 71.7 \\
        Qwen3-VL & Qwen2.5-Omni & Gemini-2.5 &  77.1 \\
        Gemini-2.5 & Gemini-2.5 & Gemini-2.5 & \textbf{83.3} \\
        \bottomrule
    \end{tabular}
    }
    \label{tab:abs_tool_model}
    \vspace{-5mm}
\end{table}

\subsection{Ablation Study}
\textbf{Tool Model Choices.}
\label{sec:abs_tool_model}
\cref{tab:abs_tool_model} delineates the impact of backbone model selection for multimodal tools on the overall performance.
Notably, the efficacy of the event model proves pivotal for the agent's reasoning capabilities. 
Given OmniAgent's substantial reliance on audio temporal grounding, Gemini-2.5 demonstrates superior proficiency in this domain. 
Conversely, previous Gemini iterations and open-source alternatives exhibit suboptimal temporal grounding. \emph{This reminds us that focusing on and enhancing the audio-visual temporal grounding capabilities of OmniLLMs is a promising direction for future research} (discussed in detail in Appendix \cref{sec:more_dis}). 
Furthermore, Qwen2.5-Omni~\cite{xu2025qwen2} suffers from performance degradation due to its inherent limitations in ASR and general audio comprehension. 
Surprisingly, employing Gemini 2.5-Flash across all tools yields better reasoning accuracy.

\begin{table}[t]
    \centering
    \caption{Ablation study on the toolset components of OmniAgent. The results demonstrate the necessity of each tool in our agent.}
    \vspace{0mm}
    \setlength{\tabcolsep}{8pt}
    \resizebox{\linewidth}{!}{
    \begin{tabular}{ccc|c}
        \toprule
        \multicolumn{3}{c|}{\textbf{Multimodal Understanding Tools}} & \textbf{Daily-Omni} \\
        \textit{Video Clip QA} & \textit{Audio QA} & \textit{Event Location} & (Avg.) \\
        \midrule
                 & \ding{51} & \ding{51} & 76.3 \\
        \ding{51} &           & \ding{51} & 80.2 \\
        \ding{51} &  &           &  77.3 \\
        \rowcolor{mycol2!20!white}
        \ding{51} & \ding{51} & \ding{51} & \textbf{82.7} \\
        \bottomrule
    \end{tabular}
    }
    \label{tab:abs:tool}
    \vspace{-5mm}
\end{table}

\textbf{Mutimodal Tools.}
\cref{tab:abs:tool} details the ablation experiments conducted to evaluate the individual contributions of various tools. 
First, Video Clip QA proves indispensable for maintaining system accuracy. In its absence, the agent reverts to repetitive reliance on Global QA, consequently failing to resolve fine-grained details. 
Furthermore, the criticality of event tools is substantiated by the data. The exclusion of Audio QA and event localization tools precipitates a substantial performance decline, thereby validating the efficacy of the proposed agentic framework and tool design.

\section{Conclusion}
We introduce \OmniAgent, a fully active perception agent tailored for omnimodal audio-visual reasoning. Operating via a recursive \emph{``Think-Act-Observe-Reflect''} loop, the system actively orchestrates tools to accumulate multimodal evidence, facilitating fine-grained comprehension progressively. Departing from conventional static workflows, we integrate a novel audio-driven event localization mechanism. This enables the model to autonomously select query-relevant information across modalities, thereby addressing the challenges of cross-modal alignment and fine-grained understanding. Experimental evaluations across diverse benchmarks demonstrate that OmniAgent significantly outperforms existing open-source and closed-source OmniLLMs.

\textbf{Discussion.} To our best knowledge, this work represents the first investigation of active perception agent technology for omnimodal audio-video understanding. 
While the current reliance on external models and extended contexts improves performance, it constrains reasoning efficiency. 
To address this, in the future, we envision training an omnimodal agentic model.
This architecture will ingest diverse modal inputs and feature tool self-calling, enabling the system to actively decide how to attend to specific audio or visual and address the bottleneck of inference cost. 
Thus, this work constitutes a pivotal bridge facilitating the advancement of omnimodal agentic algorithms and training. More discussion and insights are in Appendix~\ref{sec:more_dis}.

\clearpage
\section*{Impact Statement}
OmniAgent seeks to advance the field of machine visual and auditory understanding, equipping artificial intelligence with superior precision and flexibility for comprehending complex audio-video input. By facilitating a paradigm shift from passive reception to active exploration, this approach offers robust cross-modal, fine-grained reasoning capabilities, which hold substantial social value. However, as this technology is designed for the in-depth analysis of real-world video and audio streams, it inherently necessitates the consideration of privacy protection and data sensitivity. We acknowledge these potential ethical implications. Consequently, our evaluation is strictly confined to open-source available benchmark datasets~\cite{zhou2025daily,hong2025worldsense,li2025omnivideobench}. Furthermore, our system design prioritizes transparency and interpretability within the tool invocation and decision-making processes.

\section*{Release Notes}
\begin{itemize}
    \item \textbf{v1: Technical Report.} This version
    presents the idea, method, major experimental results, and discussions.
    \item \textbf{v2: Full Paper.} This version
    presents all experimental results, case studies, ideas, and details.
\end{itemize}

\bibliography{main}

@String(CVPR= {IEEE Conf. Comput. Vis. Pattern Recog.})

@String(ECCV= {Eur. Conf. Comput. Vis.})

@String(ICLR = {Int. Conf. Learn. Represent.})

@String(CVPR  = {CVPR})

@String(ECCV  = {ECCV})

@String(ICLR  = {ICLR})

@String(ACL = {ACL})

@String(ICML = {ICML})

@String(EMNLP = {EMNLP})

@article{li2025omnivideobench,
  title={Omnivideobench: Towards audio-visual understanding evaluation for omni mllms},
  author={Li, Caorui and Chen, Yu and Ji, Yiyan and Xu, Jin and Cui, Zhenyu and Li, Shihao and Zhang, Yuanxing and Tang, Jiafu and Song, Zhenghao and Zhang, Dingling and others},
  journal={arXiv preprint arXiv:2510.10689},
  year={2025}
}

@article{damonlpsg2024videollama2,
  title={VideoLLaMA 2: Advancing Spatial-Temporal Modeling and Audio Understanding in Video-LLMs},
  author={Cheng, Zesen and Leng, Sicong and Zhang, Hang and Xin, Yifei and Li, Xin and Chen, Guanzheng and Zhu, Yongxin and Zhang, Wenqi and Luo, Ziyang and Zhao, Deli and Bing, Lidong},
  journal={arXiv preprint arXiv:2406.07476},
  year={2024},
  url = {https://arxiv.org/abs/2406.07476}
}

@article{li2025baichuan,
  title={Baichuan-omni-1.5 technical report},
  author={Li, Yadong and Liu, Jun and Zhang, Tao and Chen, Song and Li, Tianpeng and Li, Zehuan and Liu, Lijun and Ming, Lingfeng and Dong, Guosheng and Pan, Da and others},
  journal={arXiv preprint arXiv:2501.15368},
  year={2025}
}

@software{openai2023gpt4,
  author       = {OpenAI},
  title        = {GPT-4 Model},
  year         = 2023,
  url          = {https://platform.openai.com},
  note         = {Accessed: 2023-11-08}
}

@article{cao2025xgc,
  title={Xgc-avis: Towards audio-visual content understanding with a multi-agent collaborative system},
  author={Cao, Yuqin and Min, Xiongkuo and Gao, Yixuan and Sun, Wei and Zhang, Zicheng and Han, Jinliang and Zhai, Guangtao},
  journal={arXiv preprint arXiv:2509.23251},
  year={2025}
}

@article{liu2025ola,
title={Ola: Pushing the Frontiers of Omni-Modal Language Model with Progressive Modality Alignment},
author={Liu, Zuyan and Dong, Yuhao and Wang, Jiahui and Liu, Ziwei and Hu, Winston and Lu, Jiwen and Rao, Yongming},
journal={arXiv preprint arXiv:2502.04328},
year={2025}
}

@inproceedings{lu2024unified,
  title={Unified-io 2: Scaling autoregressive multimodal models with vision language audio and action},
  author={Lu, Jiasen and Clark, Christopher and Lee, Sangho and Zhang, Zichen and Khosla, Savya and Marten, Ryan and Hoiem, Derek and Kembhavi, Aniruddha},
  booktitle=CVPR,
  year={2024}
}

@article{wang2024qwen2,
  title={Qwen2-vl: Enhancing vision-language model's perception of the world at any resolution},
  author={Wang, Peng and Bai, Shuai and Tan, Sinan and Wang, Shijie and Fan, Zhihao and Bai, Jinze and Chen, Keqin and Liu, Xuejing and Wang, Jialin and Ge, Wenbin and others},
  journal={arXiv preprint arXiv:2409.12191},
  year={2024}
}

@article{bai2025qwen2,
  title={Qwen2. 5-vl technical report},
  author={Bai, Shuai and Chen, Keqin and Liu, Xuejing and Wang, Jialin and Ge, Wenbin and Song, Sibo and Dang, Kai and Wang, Peng and Wang, Shijie and Tang, Jun and others},
  journal={arXiv preprint arXiv:2502.13923},
  year={2025}
}

@article{li2024baichuanomni,  
  title={Baichuan-Omni Technical Report},  
  author={Li, Yadong and Sun, Haoze and Lin, Mingan and Li, Tianpeng and Dong, Guosheng and Zhang, Tao and Ding, Bowen and Song, Wei and Cheng, Zhenglin and Huo, Yuqi and Chen, Song and Li, Xu and Pan, Da and Zhang, Shusen and Wu, Xin and Liang, Zheng and Liu, Jun and Zhang, Tao and Lu, Keer and Zhao, Yaqi and Shen, Yanjun and Yang, Fan and Yu, Kaicheng and Lin, Tao and Xu, Jianhua and Zhou, Zenan and Chen, Weipeng},  
  journal={arXiv preprint arXiv:2410.08565},  
  year={2024}  
}

@article{tang2025video,
  title={video-SALMONN 2: Captioning-Enhanced Audio-Visual Large Language Models},
  author={Tang, Changli and Li, Yixuan and Yang, Yudong and Zhuang, Jimin and Sun, Guangzhi and Li, Wei and Ma, Zejun and Zhang, Chao},
  journal={arXiv preprint arXiv:2506.15220},
  year={2025}
}

@misc{zhang2024videoin, title={Video Instruction Tuning With Synthetic Data}, author={Yuanhan Zhang and Jinming Wu and Wei Li and Bo Li and Zejun Ma and Ziwei Liu and Chunyuan Li}, year={2024}, eprint={2410.02713}, archivePrefix={arXiv}, primaryClass={cs.CV}, url={https://arxiv.org/abs/2410.02713}, }

@article{xu2025qwen2,
  title={Qwen2. 5-omni technical report},
  author={Xu, Jin and Guo, Zhifang and He, Jinzheng and Hu, Hangrui and He, Ting and Bai, Shuai and Chen, Keqin and Wang, Jialin and Fan, Yang and Dang, Kai and others},
  journal={arXiv preprint arXiv:2503.20215},
  year={2025}
}

@article{xu2025qwen3,
  title={Qwen3-omni technical report},
  author={Xu, Jin and Guo, Zhifang and Hu, Hangrui and Chu, Yunfei and Wang, Xiong and He, Jinzheng and Wang, Yuxuan and Shi, Xian and He, Ting and Zhu, Xinfa and others},
  journal={arXiv preprint arXiv:2509.17765},
  year={2025}
}

@article{yang2025humanomniv2,
  title={HumanOmniV2: From Understanding to Omni-Modal Reasoning with Context},
  author={Yang, Qize and Yao, Shimin and Chen, Weixuan and Fu, Shenghao and Bai, Detao and Zhao, Jiaxing and Sun, Boyuan and Yin, Bowen and Wei, Xihan and Zhou, Jingren},
  journal={arXiv preprint arXiv:2506.21277},
  year={2025}
}

@inproceedings{tao2025dycoke,
  title={Dycoke: Dynamic compression of tokens for fast video large language models},
  author={Tao, Keda and Qin, Can and You, Haoxuan and Sui, Yang and Wang, Huan},
  booktitle=CVPR,
  year={2025}
}

@article{shao2025holitom,
  title={Holitom: Holistic token merging for fast video large language models},
  author={Shao, Kele and Tao, Keda and Qin, Can and You, Haoxuan and Sui, Yang and Wang, Huan},
  journal={arXiv preprint arXiv:2505.21334},
  year={2025}
}

@inproceedings{shen2024longvu,
  title={Longvu: Spatiotemporal adaptive compression for long video-language understanding},
  author={Shen, Xiaoqian and Xiong, Yunyang and Zhao, Changsheng and Wu, Lemeng and Chen, Jun and Zhu, Chenchen and Liu, Zechun and Xiao, Fanyi and Varadarajan, Balakrishnan and Bordes, Florian and others},
  booktitle=ICML,
  year={2025}
}

@inproceedings{yang2025audio,
  title={Audio-centric video understanding benchmark without text shortcut},
  author={Yang, Yudong and Zhuang, Jimin and Sun, Guangzhi and Tang, Changli and Li, Yixuan and Li, Peihan and Jiang, Yifan and Li, Wei and Ma, Zejun and Zhang, Chao},
  booktitle=EMNLP,
  year={2025}
}

@article{ge2025arc,
  title={Arc-hunyuan-video-7b: Structured video comprehension of real-world shorts},
  author={Ge, Yuying and Ge, Yixiao and Li, Chen and Wang, Teng and Pu, Junfu and Li, Yizhuo and Qiu, Lu and Ma, Jin and Duan, Lisheng and Zuo, Xinyu and others},
  journal={arXiv preprint arXiv:2507.20939},
  year={2025}
}

@article{hong2025worldsense,
  title={Worldsense: Evaluating real-world omnimodal understanding for multimodal llms},
  author={Hong, Jack and Yan, Shilin and Cai, Jiayin and Jiang, Xiaolong and Hu, Yao and Xie, Weidi},
  journal={arXiv preprint arXiv:2502.04326},
  year={2025}
}

@article{tong2025interactiveomni,
  title={InteractiveOmni: A Unified Omni-modal Model for Audio-Visual Multi-turn Dialogue},
  author={Tong, Wenwen and Guo, Hewei and Ran, Dongchuan and Chen, Jiangnan and Lu, Jiefan and Wang, Kaibin and Li, Keqiang and Zhu, Xiaoxu and Li, Jiakui and Li, Kehan and others},
  journal={arXiv preprint arXiv:2510.13747},
  year={2025}
}

@article{xie2024mini,
  title={Mini-omni2: Towards open-source gpt-4o with vision, speech and duplex capabilities},
  author={Xie, Zhifei and Wu, Changqiao},
  journal={arXiv preprint arXiv:2410.11190},
  year={2024}
}

@inproceedings{shu2025audio,
  title={Audio-visual llm for video understanding},
  author={Shu, Fangxun and Zhang, Lei and Jiang, Hao and Xie, Cihang},
  booktitle=CVPR,
  year={2025}
}

@article{wang2025internvl3,
  title={Internvl3. 5: Advancing open-source multimodal models in versatility, reasoning, and efficiency},
  author={Wang, Weiyun and Gao, Zhangwei and Gu, Lixin and Pu, Hengjun and Cui, Long and Wei, Xingguang and Liu, Zhaoyang and Jing, Linglin and Ye, Shenglong and Shao, Jie and others},
  journal={arXiv preprint arXiv:2508.18265},
  year={2025}
}

@article{team2025kimi,
  title={Kimi-vl technical report},
  author={Team, Kimi and Du, Angang and Yin, Bohong and Xing, Bowei and Qu, Bowen and Wang, Bowen and Chen, Cheng and Zhang, Chenlin and Du, Chenzhuang and Wei, Chu and others},
  journal={arXiv preprint arXiv:2504.07491},
  year={2025}
}

@article{lin2024awq,
  title={Awq: Activation-aware weight quantization for on-device llm compression and acceleration},
  author={Lin, Ji and Tang, Jiaming and Tang, Haotian and Yang, Shang and Chen, Wei-Ming and Wang, Wei-Chen and Xiao, Guangxuan and Dang, Xingyu and Gan, Chuang and Han, Song},
  journal=NeurIPS,
  year={2024}
}

@article{sun2023simple,
  title={A simple and effective pruning approach for large language models},
  author={Sun, Mingjie and Liu, Zhuang and Bair, Anna and Kolter, J Zico},
  journal={arXiv preprint arXiv:2306.11695},
  year={2023}
}

@article{xia2023sheared,
  title={Sheared llama: Accelerating language model pre-training via structured pruning},
  author={Xia, Mengzhou and Gao, Tianyu and Zeng, Zhiyuan and Chen, Danqi},
  journal={arXiv preprint arXiv:2310.06694},
  year={2023}
}

@article{van2024gptvq,
  title={Gptvq: The blessing of dimensionality for llm quantization},
  author={Van Baalen, Mart and Kuzmin, Andrey and Koryakovskiy, Ivan and Nagel, Markus and Couperus, Peter and Bastoul, Cedric and Mahurin, Eric and Blankevoort, Tijmen and Whatmough, Paul},
  journal={arXiv preprint arXiv:2402.15319},
  year={2024}
}

@article{sun2024video,
  title={video-salmonn: Speech-enhanced audio-visual large language models},
  author={Sun, Guangzhi and Yu, Wenyi and Tang, Changli and Chen, Xianzhao and Tan, Tian and Li, Wei and Lu, Lu and Ma, Zejun and Wang, Yuxuan and Zhang, Chao},
  journal={arXiv preprint arXiv:2406.15704},
  year={2024}
}

@article{ai2025ming,
  title={Ming-Omni: A Unified Multimodal Model for Perception and Generation},
  author={AI, Inclusion and Gong, Biao and Zou, Cheng and Zheng, Chuanyang and Zhou, Chunluan and Yan, Canxiang and Jin, Chunxiang and Shen, Chunjie and Zheng, Dandan and Wang, Fudong and others},
  journal={arXiv preprint arXiv:2506.09344},
  year={2025}
}

@article{yin2025videoarm,
  title={VideoARM: Agentic Reasoning over Hierarchical Memory for Long-Form Video Understanding},
  author={Yin, Yufei and Meng, Qianke and Chen, Minghao and Ding, Jiajun and Shao, Zhenwei and Yu, Zhou},
  journal={arXiv preprint arXiv:2512.12360},
  year={2025}
}

@article{chu2023qwen,
  title={Qwen-audio: Advancing universal audio understanding via unified large-scale audio-language models},
  author={Chu, Yunfei and Xu, Jin and Zhou, Xiaohuan and Yang, Qian and Zhang, Shiliang and Yan, Zhijie and Zhou, Chang and Zhou, Jingren},
  journal={arXiv preprint arXiv:2311.07919},
  year={2023}
}

@article{zhou2025daily,
  title={Daily-Omni: Towards Audio-Visual Reasoning with Temporal Alignment across Modalities},
  author={Zhou, Ziwei and Wang, Rui and Wu, Zuxuan},
  journal={arXiv preprint arXiv:2505.17862},
  year={2025}
}

@article{bai2025qwen3vl,
  title={Qwen3-VL Technical Report},
  author={Bai, Shuai and Cai, Yuxuan and Chen, Ruizhe and Chen, Keqin and Chen, Xionghui and Cheng, Zesen and Deng, Lianghao and Ding, Wei and Gao, Chang and Ge, Chunjiang and others},
  journal={arXiv preprint arXiv:2511.21631},
  year={2025}
}

@article{ding2025kimi,
  title={Kimi-audio technical report},
  author={Ding, Ding and Ju, Zeqian and Leng, Yichong and Liu, Songxiang and Liu, Tong and Shang, Zeyu and Shen, Kai and Song, Wei and Tan, Xu and Tang, Heyi and others},
  journal={arXiv preprint arXiv:2504.18425},
  year={2025}
}

@inproceedings{zhang2024simple,
  title={A simple llm framework for long-range video question-answering},
  author={Zhang, Ce and Lu, Taixi and Islam, Md Mohaiminul and Wang, Ziyang and Yu, Shoubin and Bansal, Mohit and Bertasius, Gedas},
  booktitle=EMNLP,
  year={2024}
}

@article{wang2025videochat,
  title={VideoChat-A1: Thinking with Long Videos by Chain-of-Shot Reasoning},
  author={Wang, Zikang and Chen, Boyu and Yue, Zhengrong and Wang, Yi and Qiao, Yu and Wang, Limin and Wang, Yali},
  journal={arXiv preprint arXiv:2506.06097},
  year={2025}
}

@article{park2024too,
  title={Too many frames, not all useful: Efficient strategies for long-form video qa},
  author={Park, Jongwoo and Ranasinghe, Kanchana and Kahatapitiya, Kumara and Ryu, Wonjeong and Kim, Donghyun and Ryoo, Michael S},
  journal={arXiv preprint arXiv:2406.09396},
  year={2024}
}

@inproceedings{ma2025drvideo,
  title={Drvideo: Document retrieval based long video understanding},
  author={Ma, Ziyu and Gou, Chenhui and Shi, Hengcan and Sun, Bin and Li, Shutao and Rezatofighi, Hamid and Cai, Jianfei},
  booktitle=CVPR,
  year={2025}
}

@inproceedings{kahatapitiya2025language,
  title={Language repository for long video understanding},
  author={Kahatapitiya, Kumara and Ranasinghe, Kanchana and Park, Jongwoo and Ryoo, Michael S},
  booktitle=ACL,
  year={2025}
}

@article{kugo2025videomultiagents,
  title={VideoMultiAgents: A Multi-Agent Framework for Video Question Answering},
  author={Kugo, Noriyuki and Li, Xiang and Li, Zixin and Gupta, Ashish and Khatua, Arpandeep and Jain, Nidhish and Patel, Chaitanya and Kyuragi, Yuta and Ishii, Yasunori and Tanabiki, Masamoto and others},
  journal={arXiv preprint arXiv:2504.20091},
  year={2025}
}

@inproceedings{chowdhury2024meerkat,
  title={Meerkat: Audio-visual large language model for grounding in space and time},
  author={Chowdhury, Sanjoy and Nag, Sayan and Dasgupta, Subhrajyoti and Chen, Jun and Elhoseiny, Mohamed and Gao, Ruohan and Manocha, Dinesh},
  booktitle=ECCV,
  year={2024},
}

@article{sun2023fine,
  title={Fine-grained audio-visual joint representations for multimodal large language models},
  author={Sun, Guangzhi and Yu, Wenyi and Tang, Changli and Chen, Xianzhao and Tan, Tian and Li, Wei and Lu, Lu and Ma, Zejun and Zhang, Chao},
  journal={arXiv preprint arXiv:2310.05863},
  year={2023}
}

@article{fan2025fine,
  title={Fine-Grained Audio--Visual Event Localization},
  author={Fan, Baoyu and Liu, Lu and Li, Xiaochuan and Zhang, Runze and Jin, Liang and Zhang, Jin},
  journal={IEEE Transactions on Neural Networks and Learning Systems},
  year={2025},
  publisher={IEEE}
}

@article{galougah2025aura,
  title={AURA: A Fine-Grained Benchmark and Decomposed Metric for Audio-Visual Reasoning},
  author={Galougah, Siminfar Samakoush and Raj, Rishie and Chowdhury, Sanjoy and Nag, Sayan and Duraiswami, Ramani},
  journal={arXiv preprint arXiv:2508.07470},
  year={2025}
}

@inproceedings{jiang2025specific,
  title={From specific-MLLMs to OMNI-MLLMs: a survey on MLLMs aligned with multi-modalities},
  author={Jiang, Shixin and Liang, Jiafeng and Wang, Jiyuan and Dong, Xuan and Chang, Heng and Yu, Weijiang and Du, Jinhua and Liu, Ming and Qin, Bing},
  booktitle={Findings of ACL},
  year={2025}
}

@article{liu2025videomind,
  title={VideoMind: A Chain-of-LoRA Agent for Long Video Reasoning},
  author={Liu, Ye and Lin, Kevin Qinghong and Chen, Chang Wen and Shou, Mike Zheng},
  journal={arXiv preprint arXiv:2503.13444},
  year={2025}
}

@inproceedings{min2024morevqa,
  title={Morevqa: Exploring modular reasoning models for video question answering},
  author={Min, Juhong and Buch, Shyamal and Nagrani, Arsha and Cho, Minsu and Schmid, Cordelia},
  booktitle={Proceedings of the IEEE/CVF Conference on Computer Vision and Pattern Recognition},
  pages={13235--13245},
  year={2024}
}

@inproceedings{guo2025aligned,
  title={Aligned Better, Listen Better for Audio-Visual Large Language Models},
  author={Guo, Yuxin and Ma, Shuailei and Ma, Shijie and Bao, Xiaoyi and Xie, Chen-Wei and Zheng, Kecheng and Weng, Tingyu and Sun, Siyang and Zheng, Yun and Zou, Wei},
  booktitle={ICLR},
  year={2025}
}

@article{chowdhury2025magnet,
  title={MAGNET: A Multi-agent Framework for Finding Audio-Visual Needles by Reasoning over Multi-Video Haystacks},
  author={Chowdhury, Sanjoy and Elmoghany, Mohamed and Abeysinghe, Yohan and Fei, Junjie and Nag, Sayan and Khan, Salman and Elhoseiny, Mohamed and Manocha, Dinesh},
  journal={arXiv preprint arXiv:2506.07016},
  year={2025}
}

@article{zhang2024omagent,
  title={OmAgent: A Multi-modal Agent Framework for Complex Video Understanding with Task Divide-and-Conquer},
  author={Zhang, Lu and Zhao, Tiancheng and Ying, Heting and Ma, Yibo and Lee, Kyusong},
  journal={arXiv preprint arXiv:2406.16620},
  year={2024}
}

@article{yang2025streamagent,
  title={Streamagent: Towards anticipatory agents for streaming video understanding},
  author={Yang, Haolin and Tang, Feilong and Zhao, Lingxiao and An, Xiang and Hu, Ming and Li, Huifa and Zhuang, Xinlin and Lu, Yifan and Zhang, Xiaofeng and Swikir, Abdalla and others},
  journal={arXiv preprint arXiv:2508.01875},
  year={2025}
}

@article{gao2025agentic,
  title={Agentic Video Intelligence: A Flexible Framework for Advanced Video Exploration and Understanding},
  author={Gao, Hong and Bao, Yiming and Tu, Xuezhen and Xu, Yutong and Jin, Yue and Mu, Yiyang and Zhong, Bin and Yue, Linan and Zhang, Min-Ling},
  journal={arXiv preprint arXiv:2511.14446},
  year={2025}
}

@article{yuan2025videodeepresearch,
  title={VideoDeepResearch: Long Video Understanding With Agentic Tool Using},
  author={Yuan, Huaying and Liu, Zheng and Zhou, Junjie and Wen, Ji-Rong and Dou, Zhicheng},
  journal={arXiv preprint arXiv:2506.10821},
  year={2025}
}

@article{zhu2025active,
  title={Active-O3: Empowering Multimodal Large Language Models with Active Perception via GRPO},
  author={Zhu, Muzhi and Zhong, Hao and Zhao, Canyu and Du, Zongze and Huang, Zheng and Liu, Mingyu and Chen, Hao and Zou, Cheng and Chen, Jingdong and Yang, Ming and others},
  journal={arXiv preprint arXiv:2505.21457},
  year={2025}
}

@article{feng2025can,
  title={Can MLLMs Guide Me Home? A Benchmark Study on Fine-Grained Visual Reasoning from Transit Maps},
  author={Feng, Sicheng and Wang, Song and Ouyang, Shuyi and Kong, Lingdong and Song, Zikai and Zhu, Jianke and Wang, Huan and Wang, Xinchao},
  journal={arXiv preprint arXiv:2505.18675},
  year={2025}
}

@article{feng2025rewardmap,
  title={RewardMap: Tackling Sparse Rewards in Fine-grained Visual Reasoning via Multi-Stage Reinforcement Learning},
  author={Feng, Sicheng and Tuo, Kaiwen and Wang, Song and Kong, Lingdong and Zhu, Jianke and Wang, Huan},
  journal={arXiv preprint arXiv:2510.02240},
  year={2025}
}

@article{feng2025efficient,
  title={Efficient reasoning models: A survey},
  author={Feng, Sicheng and Fang, Gongfan and Ma, Xinyin and Wang, Xinchao},
  journal={arXiv preprint arXiv:2504.10903},
  year={2025}
}

@article{team2023gemini,
  title={Gemini: a family of highly capable multimodal models},
  author={Team, Gemini and Anil, Rohan and Borgeaud, Sebastian and Alayrac, Jean-Baptiste and Yu, Jiahui and Soricut, Radu and Schalkwyk, Johan and Dai, Andrew M and Hauth, Anja and Millican, Katie and others},
  journal={arXiv preprint arXiv:2312.11805},
  year={2023}
}

@article{team2024gemini,
  title={Gemini 1.5: Unlocking multimodal understanding across millions of tokens of context},
  author={Team, Gemini and Georgiev, Petko and Lei, Ving Ian and Burnell, Ryan and Bai, Libin and Gulati, Anmol and Tanzer, Garrett and Vincent, Damien and Pan, Zhufeng and Wang, Shibo and others},
  journal={arXiv preprint arXiv:2403.05530},
  year={2024}
}

@article{chen2025chronusomni,
  title={ChronusOmni: Improving Time Awareness of Omni Large Language Models},
  author={Chen, Yijing and Wu, Yihan and Guan, Kaisi and Ren, Yuchen and Wang, Yuyue and Song, Ruihua and Ru, Liyun},
  journal={arXiv preprint arXiv:2512.09841},
  year={2025}
}

@article{wu2025survey,
  title={A survey on video temporal grounding with multimodal large language model},
  author={Wu, Jianlong and Liu, Wei and Liu, Ye and Liu, Meng and Nie, Liqiang and Lin, Zhouchen and Chen, Chang Wen},
  journal={IEEE Transactions on Pattern Analysis and Machine Intelligence},
  year={2025},
  publisher={IEEE}
}

@article{jeoung2024adaptive,
  title={Adaptive Video Understanding Agent: Enhancing efficiency with dynamic frame sampling and feedback-driven reasoning},
  author={Jeoung, Sullam and Huybrechts, Goeric and Ganesh, Bhavana and Galstyan, Aram and Bodapati, Sravan},
  journal={arXiv preprint arXiv:2410.20252},
  year={2024}
}

@inproceedings{yao2022react,
  title={React: Synergizing reasoning and acting in language models},
  author={Yao, Shunyu and Zhao, Jeffrey and Yu, Dian and Du, Nan and Shafran, Izhak and Narasimhan, Karthik R and Cao, Yuan},
  booktitle={The eleventh international conference on learning representations},
  year={2022}
}

@inproceedings{fan2024videoagent,
  title={Videoagent: A memory-augmented multimodal agent for video understanding},
  author={Fan, Yue and Ma, Xiaojian and Wu, Rujie and Du, Yuntao and Li, Jiaqi and Gao, Zhi and Li, Qing},
  booktitle=ECCV,
  year={2024},
}

@inproceedings{wang2024videoagent,
  title={Videoagent: Long-form video understanding with large language model as agent},
  author={Wang, Xiaohan and Zhang, Yuhui and Zohar, Orr and Yeung-Levy, Serena},
  booktitle=ECCV,
  year={2024},
}

@article{zhang2025deep,
  title={Deep Video Discovery: Agentic Search with Tool Use for Long-form Video Understanding},
  author={Zhang, Xiaoyi and Jia, Zhaoyang and Guo, Zongyu and Li, Jiahao and Li, Bin and Li, Houqiang and Lu, Yan},
  journal={arXiv preprint arXiv:2505.18079},
  year={2025}
}

@inproceedings{wang2025videotree,
  title={Videotree: Adaptive tree-based video representation for llm reasoning on long videos},
  author={Wang, Ziyang and Yu, Shoubin and Stengel-Eskin, Elias and Yoon, Jaehong and Cheng, Feng and Bertasius, Gedas and Bansal, Mohit},
  booktitle=CVPR,
  year={2025}
}

@inproceedings{yang2025vca,
  title={Vca: Video curious agent for long video understanding},
  author={Yang, Zeyuan and Chen, Delin and Yu, Xueyang and Shen, Maohao and Gan, Chuang},
  booktitle=CVPR,
  year={2025}
}

@article{pang2025mr,
  title={Mr. video:" mapreduce" is the principle for long video understanding},
  author={Pang, Ziqi and Wang, Yu-Xiong},
  journal={arXiv preprint arXiv:2504.16082},
  year={2025}
}

@article{wang2025active,
  title={Active Video Perception: Iterative Evidence Seeking for Agentic Long Video Understanding},
  author={Wang, Ziyang and Zhou, Honglu and Wang, Shijie and Li, Junnan and Xiong, Caiming and Savarese, Silvio and Bansal, Mohit and Ryoo, Michael S and Niebles, Juan Carlos},
  journal={arXiv preprint arXiv:2512.05774},
  year={2025}
}

@article{tao2025omnizip,
  title={OmniZip: Audio-Guided Dynamic Token Compression for Fast Omnimodal Large Language Models},
  author={Tao, Keda and Shao, Kele and Yu, Bohan and Wang, Weiqiang and Wang, Huan and others},
  journal={arXiv preprint arXiv:2511.14582},
  year={2025}
}

@article{comanici2025gemini,
  title={Gemini 2.5: Pushing the frontier with advanced reasoning, multimodality, long context, and next generation agentic capabilities},
  author={Comanici, Gheorghe and Bieber, Eric and Schaekermann, Mike and Pasupat, Ice and Sachdeva, Noveen and Dhillon, Inderjit and Blistein, Marcel and Ram, Ori and Zhang, Dan and Rosen, Evan and others},
  journal={arXiv preprint arXiv:2507.06261},
  year={2025}
}

@inproceedings{han2016deep,
  title={Deep compression: Compressing deep neural network with pruning, trained quantization and huffman coding},
  author={Han, Song and Mao, Huizi and Dally, William J},
  booktitle=ICLR,
  year={2016}
}

@article{zhu2025obs,
  title={OBS-Diff: Accurate Pruning For Diffusion Models in One-Shot},
  author={Zhu, Junhan and Wang, Hesong and Su, Mingluo and Wang, Zefang and Wang, Huan},
  journal={arXiv preprint arXiv:2510.06751},
  year={2025}
}

@article{chu2023mobilevlm,
  title={Mobilevlm: A fast, strong and open vision language assistant for mobile devices},
  author={Chu, Xiangxiang and Qiao, Limeng and Lin, Xinyang and Xu, Shuang and Yang, Yang and Hu, Yiming and Wei, Fei and Zhang, Xinyu and Zhang, Bo and Wei, Xiaolin and others},
  journal={arXiv preprint arXiv:2312.16886},
  year={2023}
}

@article{du2025heads,
  title={Which Heads Matter for Reasoning? RL-Guided KV Cache Compression},
  author={Du, Wenjie and Jiang, Li and Tao, Keda and Liu, Xue and Wang, Huan},
  journal={arXiv preprint arXiv:2510.08525},
  year={2025}
}

@article{xie2025caption,
  title={Caption Assisted Multimodal Large Language Model for Video Moment Retrieval},
  author={Xie, Peiyu and Li, Jinxing and Lu, Guangming and Xu, Yong and Zhang, David},
  journal={IEEE Transactions on Image Processing},
  year={2025},
  publisher={IEEE}
}

@inproceedings{tian2025ego,
  title={Ego-R1: Chain-of-Tool-Thought for Ultra-Long Egocentric Video Reasoning},
  author={Tian, Shulin and Wang, Ruiqi and Guo, Hongming and Wu, Penghao and Dong, Yuhao and Wang, Xiuying and Yang, Jingkang and Zhang, Hao and Zhu, Hongyuan and Liu, Ziwei},
  booktitle=NeurIPS,
  year={2025}
}

@article{token_compression_survey,
  title={When Tokens Talk Too Much: A Survey of Multimodal Long-Context Token Compression across Images, Videos, and Audios},
  author={Shao, Kele and Tao, Keda and Zhang, Kejia and Feng, Sicheng and Cai, Mu and Shang, Yuzhang and You, Haoxuan and Qin, Can and Sui, Yang and Wang, Huan},
  journal={arXiv preprint arXiv:2507.20198},
  year={2025}
}

@article{ding2025videozoomer,
  title={VideoZoomer: Reinforcement-Learned Temporal Focusing for Long Video Reasoning},
  author={Ding, Yang and Zhang, Yizhen and Lai, Xin and Chu, Ruihang and Yang, Yujiu},
  journal={arXiv preprint arXiv:2512.22315},
  year={2025}
}

@article{wang2025time,
  title={Time-R1: Post-Training Large Vision Language Model for Temporal Video Grounding},
  author={Wang, Ye and Wang, Ziheng and Xu, Boshen and Du, Yang and Lin, Kejun and Xiao, Zihan and Yue, Zihao and Ju, Jianzhong and Zhang, Liang and Yang, Dingyi and others},
  journal={arXiv preprint arXiv:2503.13377},
  year={2025}
}

@article{zhang2025timelens,
  title={TimeLens: Rethinking Video Temporal Grounding with Multimodal LLMs},
  author={Zhang, Jun and Wang, Teng and Ge, Yuying and Ge, Yixiao and Li, Xinhao and Shan, Ying and Wang, Limin},
  journal={arXiv preprint arXiv:2512.14698},
  year={2025}
}

@article{team2025longcat,
  title={Longcat-flash-omni technical report},
  author={Team, Meituan LongCat and Wang, Bairui and Xiao, Bin and Zhang, Bo and Rong, Bolin and Chen, Borun and Wan, Chang and Zhang, Chao and Huang, Chen and Chen, Chen and others},
  journal={arXiv preprint arXiv:2511.00279},
  year={2025}
}

@article{zeng2024timesuite,
  title={Timesuite: Improving mllms for long video understanding via grounded tuning},
  author={Zeng, Xiangyu and Li, Kunchang and Wang, Chenting and Li, Xinhao and Jiang, Tianxiang and Yan, Ziang and Li, Songze and Shi, Yansong and Yue, Zhengrong and Wang, Yi and others},
  journal={arXiv preprint arXiv:2410.19702},
  year={2024}
}

@inproceedings{qu2024chatvtg,
  title={Chatvtg: Video temporal grounding via chat with video dialogue large language models},
  author={Qu, Mengxue and Chen, Xiaodong and Liu, Wu and Li, Alicia and Zhao, Yao},
  booktitle={CVPR},
  year={2024}
}

@inproceedings{huang2024vtimellm,
  title={Vtimellm: Empower llm to grasp video moments},
  author={Huang, Bin and Wang, Xin and Chen, Hong and Song, Zihan and Zhu, Wenwu},
  booktitle={CVPR},
  year={2024}
}
\bibliographystyle{icml2026}

\newpage
\appendix
\onecolumn

\begin{mytocbox}
\setlength{\parskip}{0.4em} 
    \noindent
    \hypersetup{linkcolor=brown!60!black}
    
    \textbf{A.} \hyperref[sec:appendix_implementation]{More Implementation Details} \dotfill P\pageref{sec:appendix_implementation} 
    
    \textbf{B.} \hyperref[sec:appendix_behavior]{Behavior Analyses} \dotfill P\pageref{sec:appendix_behavior} 
    
    \hspace*{1.5em} \textbf{B.1} \hyperref[sec:appendix_behavior_benchmark]{Explainability of Modal Tendency Reasoning} \dotfill P\pageref{sec:appendix_behavior_benchmark} 

    \hspace*{1.5em} \textbf{B.2} \hyperref[sec:appendix_behavior_benchmark_analyses]{Reasoning Behavior Analyses in Different Benchmarks} \dotfill P\pageref{sec:appendix_behavior_benchmark_analyses} 
        
    \textbf{C.} \hyperref[sec:appendix_efficiency]{Efficiency Analyses} \dotfill P\pageref{sec:appendix_efficiency}

    \textbf{D.} \hyperref[sec:case_study]{Case Study} \dotfill P\pageref{sec:case_study}

    \hspace*{1.5em} \textbf{D.1} \hyperref[sec:case_study_normal]{Audio-Video Reasoning Cases} \dotfill P\pageref{sec:case_study_normal} 

    \hspace*{1.5em} \textbf{D.2} \hyperref[sec:case_study_uni]{Unimodal-based Reasoning Cases} \dotfill P\pageref{sec:case_study_uni} 

    \hspace*{1.5em} \textbf{D.3} \hyperref[sec:case_study_fail]{Analyses of Failure Cases} \dotfill P\pageref{sec:case_study_fail} 

    \textbf{E.} \hyperref[sec:limitation]{Future Work} \dotfill P\pageref{sec:limitation}
    
    \textbf{F.} \hyperref[sec:more_dis]{\textcolor{MacaronGold}{\faStar} Insights and More Discussion} \dotfill P\pageref{sec:more_dis}
    
    \textbf{G.} \hyperref[sec:prompt]{Prompts of OmniAgent} \dotfill P\pageref{sec:prompt}

    \textbf{H.} \hyperref[sec:prompt-tool]{Prompts and Showcase of Toolset} \dotfill P\pageref{sec:prompt-tool}
\end{mytocbox}

\vspace{-1mm}
\section{More Implementation Details.}
\label{sec:appendix_implementation}
OmniAgent operates within an agentic, active reasoning paradigm. 
We employ the official APIs for all utilized tool models and the OpenAI model series. 
Conversely, for comparative baselines, we rely primarily on results reported by established benchmarks.
When the reasoning process exceeds the maximum step limit, we leverage OpenAI o3 to synthesize answers based on accumulated information and evidence. 
To mitigate the impact of network fluctuations and latency, any queries interrupted by connectivity issues were re-evaluated in a final testing phase.
For the DVD~\cite{zhang2025deep}, we used the official code for evaluation and set the FPS to 5 for fair comparison.
For our tools $\mathcal{T}_{\text{VGA}}$ and $\mathcal{T}_{\text{VCA}}$, we set the video sampling FPS to 2 and 5, respectively.

For DVD, due to the high API costs that we cannot afford for longer videos, we only conducted the test on Daily-Omni.
For the Daily-Omni agent~\cite{zhou2025daily}, we adopt the official performance metrics reported in their paper. 
Due to the agent's exclusive reliance on the segmented video input configuration inherent to the Daily-Omni benchmark, cross-evaluation on other benchmarks was not feasible. 
Regarding tool integration, we incorporated a supplementary video metadata tool that enables the model to autonomously retrieve essential video attributes, including duration and frame rate (FPS), thereby providing foundational data to support subsequent tool invocations. 

\textbf{Cost.} We measure the average API cost of our OmniAgent in three benchmarks. As the length of the video and the complexity of the query change, the total cost will also change accordingly. Thus, OmniAgent incurs a cost of \$0.05-\$0.11 per question with OpenAI, Qwen, and Gemini API.

\textbf{Prompts.} We show the prompts used by different tools and models within the proposed OmniAgent: (1) system prompts and user prompts for the agent (\cref{sec:prompt}); (2) prompts for the video tools (\cref{sec:prompt-tool_video_1,sec:prompt-tool_video_2}); (3) prompts for the audio tools (\cref{sec:prompt-tool_audio_1,sec:prompt-tool_audio_2,sec:prompt-tool_audio_3}); (4) prompts for the event tools (\cref{sec:prompt-tool_event_1,sec:prompt-tool_event_2}).

\section{Behavior Analyses}
\label{sec:appendix_behavior}

\subsection{Explainability of Modal Tendency Reasoning}
\label{sec:appendix_behavior_benchmark}

\begin{wrapfigure}{r}{0.43\textwidth}
\vspace{-7mm}
\centering
\captionsetup{font={small}, skip=0pt}
\includegraphics[width=1\linewidth]{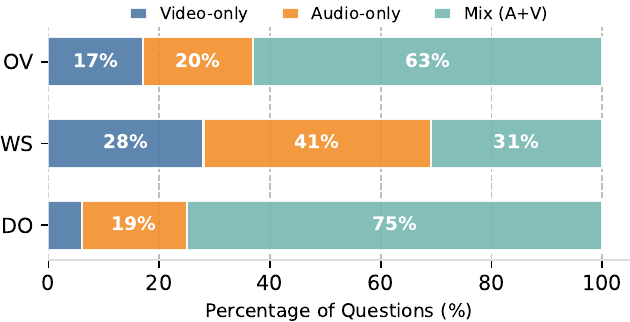}
\vspace{-3mm}
\caption{Focusing exclusively on the subset of queries correctly resolved by OmniAgent, we quantified the distribution of modal tools utilized during response generation.}
\vspace{-4mm}
\label{tab：appendix_model_ten}
\end{wrapfigure}

To analyze the performance of OmniAgent across diverse benchmarks: Daily-Omni (DO)~\cite{zhou2025daily}, WorldSense (WS)~\cite{hong2025worldsense}, and OmniVideoBench (OV)~\cite{li2025omnivideobench}, we quantified the modality distribution of the tools utilized to derive answers. 
Our analysis focused exclusively on QA pairs correctly resolved by the agent. 
As illustrated in \cref{tab：appendix_model_ten}, we categorized these instances into three primary classes based on input requirements: visual-only (video perception tools), audio-only (audio perception tools and event perception tools), and mix.
We observe that OmniAgent can obtain a correct answer, depending on the usage of unimodal tools, on some questions in the WorldSense benchmark, demonstrating the generalization ability of our agent (for more case studies, see \cref{sec:case_study}).

This indicates that during tool invocation, OmniAgent effectively infers the necessary context from the query and retrieves information aligned with the specific required modality. 
Furthermore, this demonstrates that our proposed algorithm is not strictly audio-dependent but rather possesses the capability to make autonomous, query-driven decisions.
Driven by the specialized design of the Daily-Omni and OmniVideoBench, the agent is typically required to conduct a comprehensive analysis of both modalities before formulating a response.
Therefore, OmniAgent has wide adaptability for omnimodal understanding. 
This also reflects that active perception can concentrate information in the required modalities.

\subsection{Reasoning Behavior Analyses in Different Benchmarks}
\label{sec:appendix_behavior_benchmark_analyses}
\begin{figure*}[h]
    \centering
    \includegraphics[width=\linewidth]{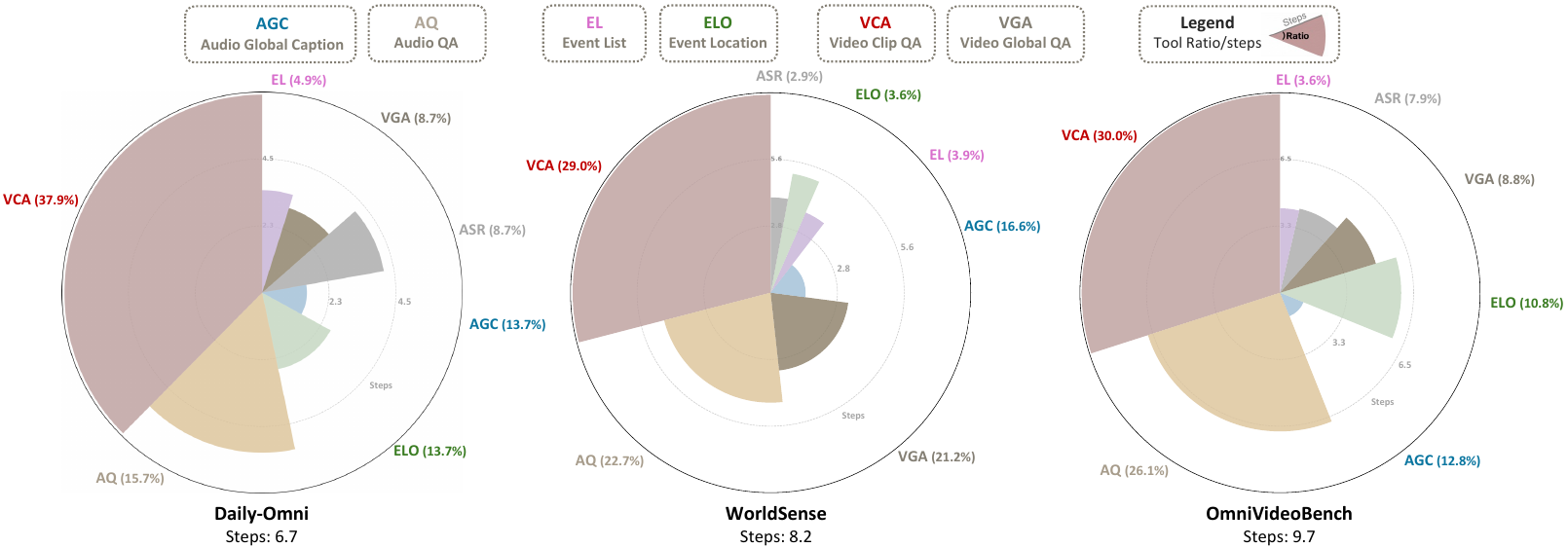}
    \caption{Analysis of the behavior of OmniAgent with OpenAI o3 in three different benchmarks.}
    \label{fig:analysis_tool_benchmark}
    \vspace{-3mm}
\end{figure*}

Consistent with the methodology outlined in \cref{sec:main_analyses_llm}, OmniAgent adheres to a coarse-to-fine reasoning paradigm to achieve fine-grained understanding. 
\cref{fig:analysis_tool_benchmark} presents the statistical analysis of inference behaviors across diverse benchmarks.
Notably, while increased video duration correlates with a higher number of iterative inference steps for information acquisition, the overall distribution of tool usage remains stable. 
Regarding WorldSense, the dataset is characterized by a high prevalence of distinctly vision-based and audio-based questions. 
For vision-centric queries, the agent typically employs a "global-to-local" strategy (\cref{fig:case_study_only_video}), leading to increased invocation frequency for Video Global QA. 
Conversely, OmniVideoBench exhibits a balanced distribution between audio-based localization and coarse-grained video localization.

\section{Efficiency Analyses}
\label{sec:appendix_efficiency}
\begin{wraptable}{r}{0.45\linewidth}
    \centering
    \vspace{-3.5mm}
    \caption{Comparison of token consumption and latency with DVD in the Daily-Omni benchmark. For a fair comparison, we use Qwen3-VL for the caption generation in DVD. }
    \vspace{-2mm}
    \resizebox{\linewidth}{!}{%
    \begin{tabular}{l|cc}
        \toprule
        \multirow{2}{*}{\textbf{Method}} & \textbf{Input Token Cost} & \textbf{Average Latency} \\
         & Visual Only & On Daily-Omni Benchmark \\
        \midrule
        DVD & 18.6k & 104s \\
        OmniAgent (Ours) & 8.3k  & 71s \\
        \bottomrule
    \end{tabular}
    }
    \label{tab:token_consumption}
    \vspace{-3mm}
\end{wraptable}
As detailed in \cref{tab:token_consumption}, a comparison with the DVD baseline~\cite {zhang2025deep} demonstrates that our method substantially reduces visual token redundancy. Moreover, despite the computational overhead incurred by processing audio signals, OmniAgent achieves a significant reduction in inference latency. 
Due to the prohibitive computational cost incurred by DVD on the remaining benchmarks, we conducted our evaluation on a randomly sampled subset of queries. Notably, in scenarios involving extended video durations, our approach reduces the aggregate API cost to approximately \textbf{10\%} of that required by DVD.
Notably, our design prioritizes solving the cross-modal alignment and fine-grained understanding over inference speed, and this work proposes a new paradigm for agent-based algorithms to solve audio-video understanding. 
We will work on more efficient improvements later.

\clearpage
\section{Case Study}
\label{sec:case_study}

\subsection{Audio-Video Reasoning Cases}
\label{sec:case_study_normal}

\begin{figure*}[t]
    \centering
    \includegraphics[width=1\linewidth]{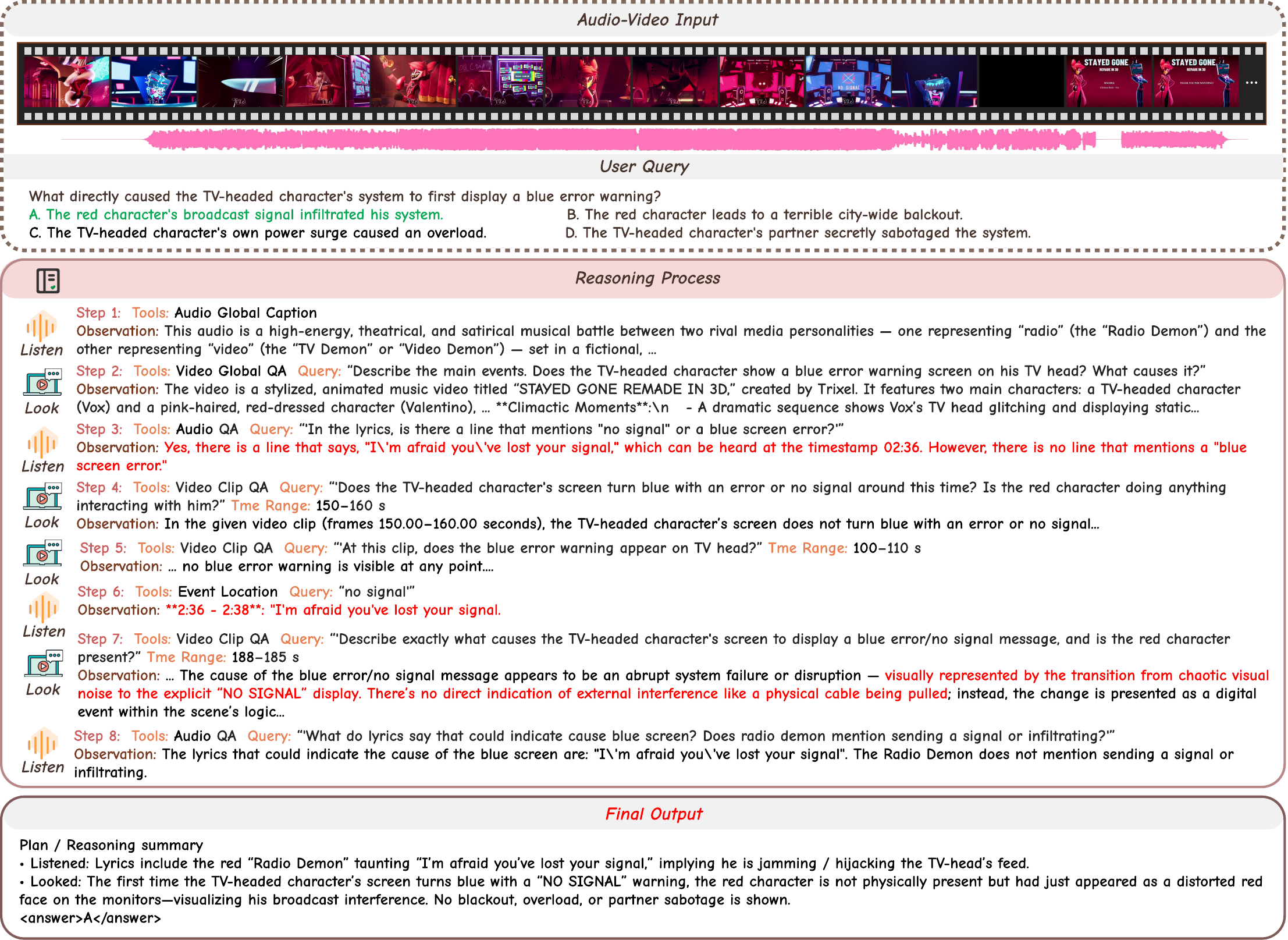}
    \caption{\textbf{Complex reasoning case in long video.} For complex problems, the intelligent agent needs to conduct repeated iterations to find key evidence and information. Moreover, we have discovered that audio plays a crucial role throughout the reasoning process.}
    \label{fig:case_study_long}
    \vspace{-4mm}
\end{figure*}

\begin{figure*}[h]
    \centering
    \includegraphics[width=1\linewidth]{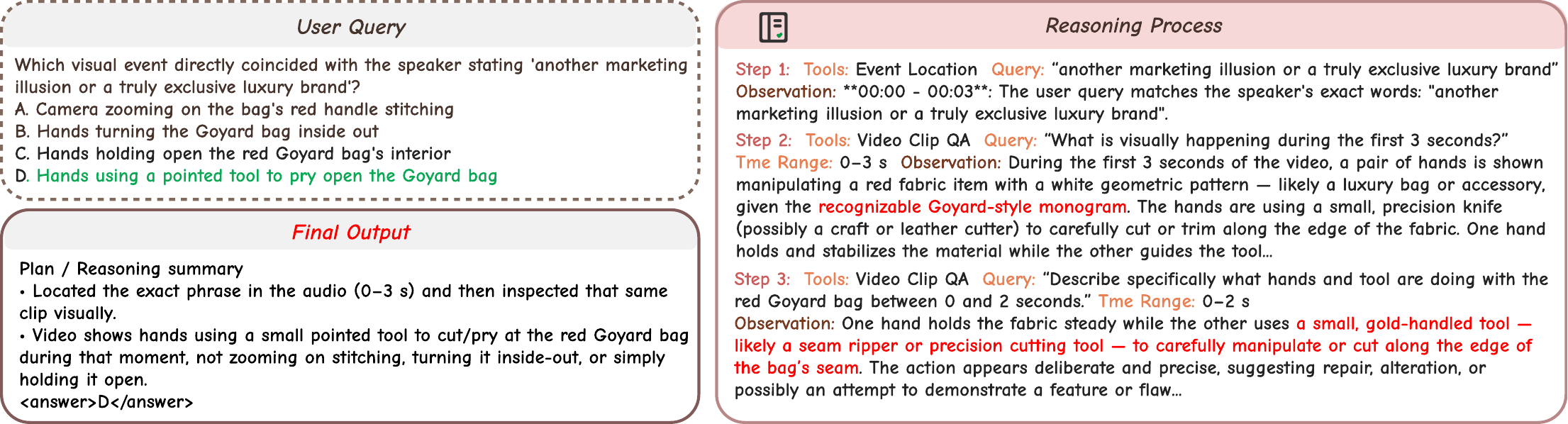}
    \caption{\textbf{Basic reasoning case in short video.} In tasks involving fundamental audio-visual event understanding, OmniAgent demonstrates high proficiency, deriving accurate answers within 2-3 reasoning steps. Specifically, the agent extracts temporal cues from the audio, subsequently leveraging this temporal grounding to align and analyze the corresponding visual information strictly.}
    \label{fig:case_study_base}
\end{figure*}

\cref{fig:case_study_base,fig:case_study_long} visualize the active decision-making and tool invocation mechanisms of OmniAgent when processing audio-visual queries. 
\cref{fig:case_study_base} depicts the reasoning workflow for fundamental questions on short video clips. 
In this scenario, the agent leverages audio cues to temporally localize events before subsequently analyzing the corresponding visual information. 
In contrast, \cref{fig:case_study_long} demonstrates a more complex reasoning scenario. 
The model initially establishes a global audio-visual context, performs joint audio-visual localization for the key cue ``no signal'', and integrates specific QA tools to extract decisive evidence. 
This exemplifies the adaptive operational logic of OmniAgent when addressing complex queries, highlighting the efficacy of audio-guided event localization.

\subsection{Unimodal-based Reasoning Cases}
\label{sec:case_study_uni}

\begin{figure*}[t]
    \centering
    \includegraphics[width=1\linewidth]{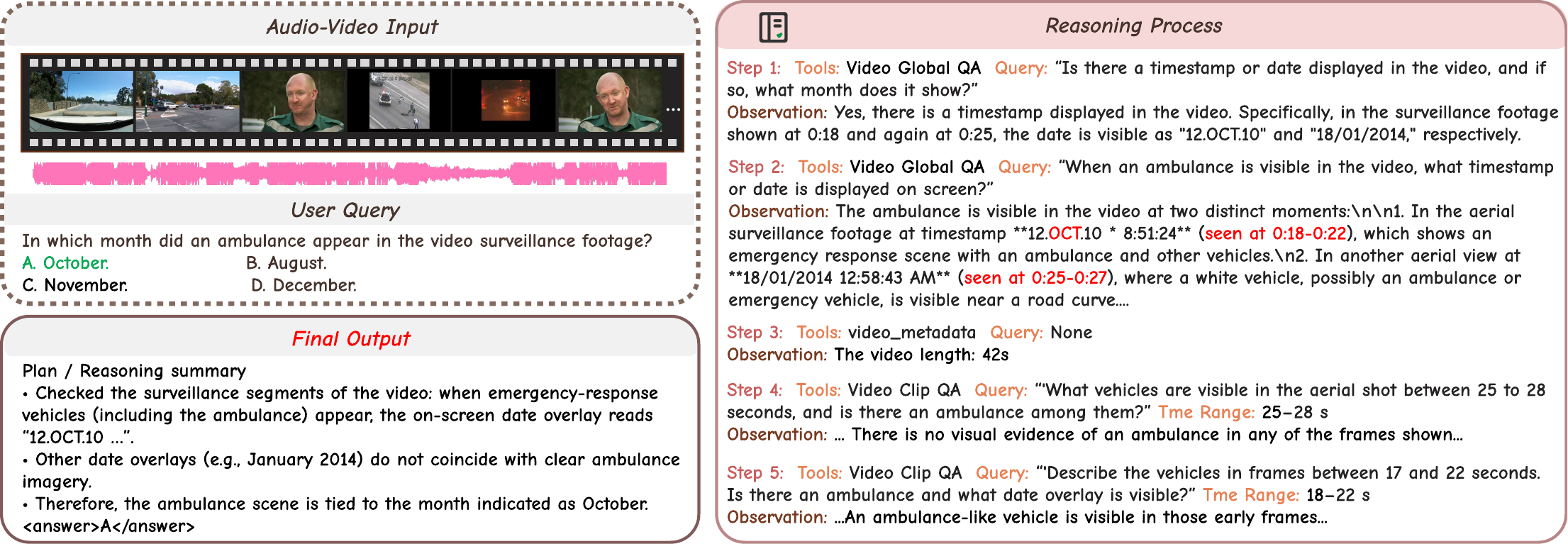}
    \caption{\textbf{Video-based reasoning process case.} For queries resolvable solely via visual information, OmniAgent autonomously elects to utilize video tools, adhering to a reasoning trajectory from coarse to fine. Typically, the agent initiates the process with Global QA for context retrieval and localization, subsequently employing Clip QA to verify fine-grained details. This validates that our framework is not rigidly constrained by audio dependency; instead, it employs adaptive, context-aware reasoning.}
    \label{fig:case_study_only_video}

\end{figure*}

\begin{figure*}[t!]
    \centering
    \includegraphics[width=1\linewidth]{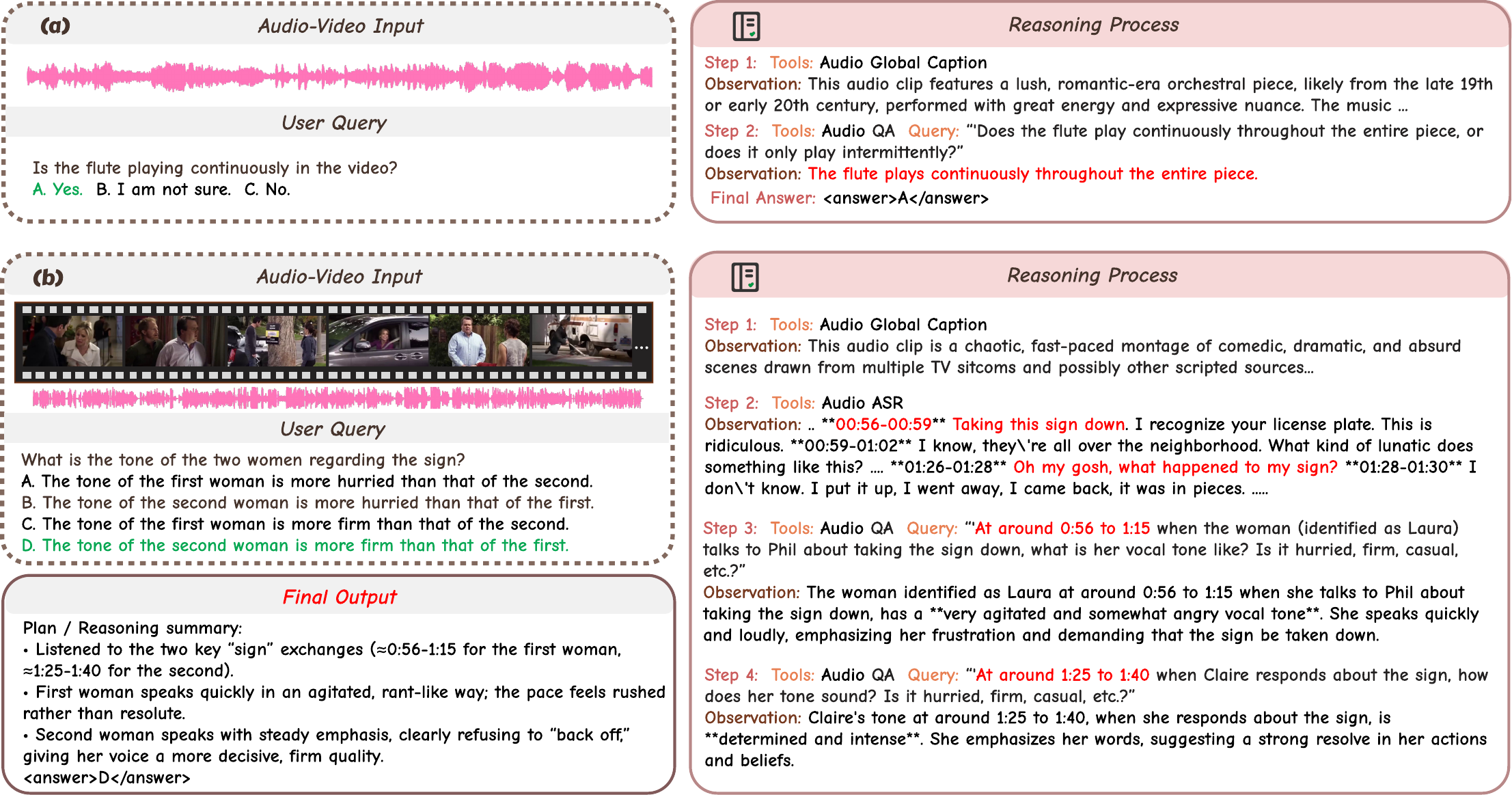}
    \caption{\textbf{Audio-based reasoning process case.} Regarding inquiries specifically targeting audio information within the video, OmniAgent strategically prioritizes audio tools, thereby avoiding superfluous visual processing and reducing information redundancy. For fundamental queries (a), the agent directly leverages audio-based interrogation to derive answers. In complex scenarios (b), the agent similarly adheres to a coarse-to-fine reasoning paradigm. As illustrated in the figure, the agent initially employs the ASR tool to transcribe the speaker's dialogue. It then analyzes this transcript relative to the query to isolate the relevant temporal segment, ultimately leveraging the Audio QA tool to verify fine-grained details, such as the speaker's emotional state. This workflow further validates the efficacy and logical soundness of our architectural design.}
    \label{fig:case_study_only_audio}
    \vspace{-1mm}
\end{figure*}

As detailed in \cref{sec:appendix_behavior_benchmark}, for tasks exhibiting a unimodal bias, such as those exclusively evaluating visual or audio comprehension, the agent invokes decision-making tools actively. 
Crucially, \emph{OmniAgent does \textbf{not enforce strict} adherence to a specific video or audio modality; rather, modality selection is dynamically determined by the agent's internal reasoning.} 
As illustrated in \cref{fig:case_study_only_video}, when addressing visual queries, OmniAgent adopts a coarse-to-fine processing strategy, aligning with our design objective of facilitating fine-grained understanding. 
This demonstrates that while audio is a powerful cue for localization, it is not forcibly integrated into scenarios where it is extraneous. 
Consequently, this confirms the capability of our agent to locate events solely via visual cues. 
Conversely, \cref{fig:case_study_only_audio} depicts a scenario where the solution is derived exclusively from audio information.

\begin{figure*}[t!]
    \centering
    \includegraphics[width=1\linewidth]{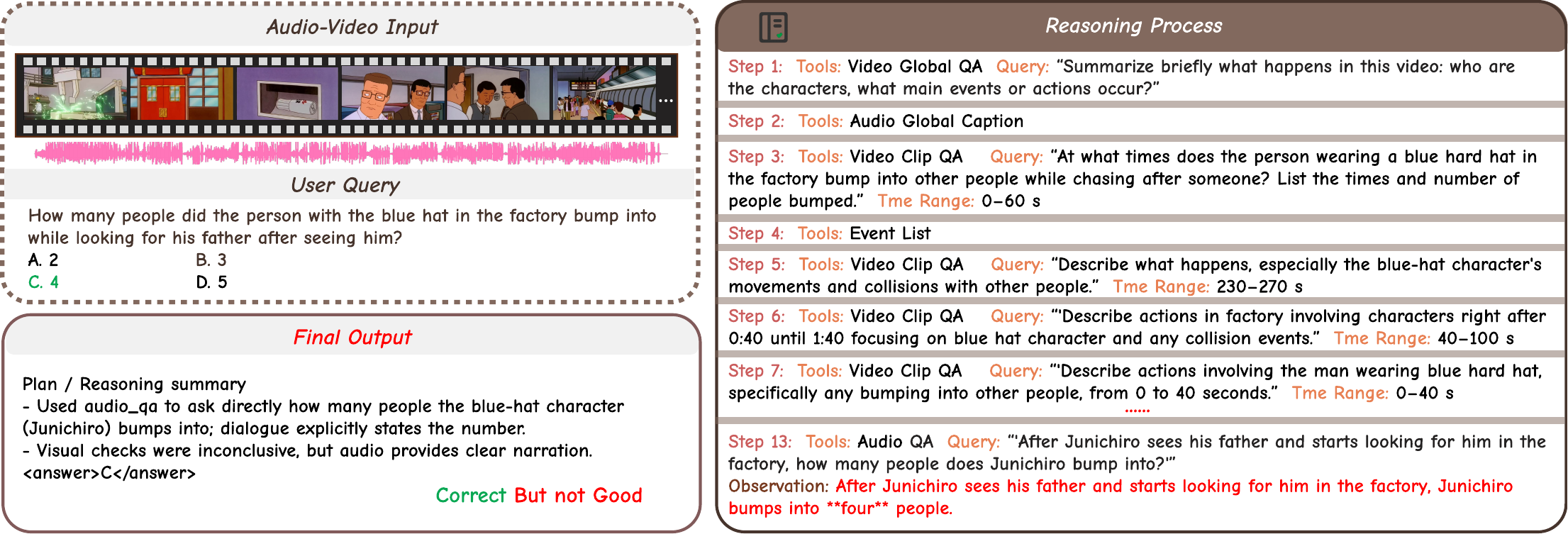}
    \caption{\textbf{Failure case analyses.} When addressing complex queries, insufficient information retrieval and the propagation of erroneous reasoning from preceding steps can cause the agent to become entrapped in a recursive cycle of redundant Video Clip QA invocations.}
    \label{fig:case_study_fail_1}
    \vspace{-3mm}
\end{figure*}

\subsection{Analyses of Failure Cases}
\label{sec:case_study_fail}
To provide a balanced analysis of agent behavior, we also present a case study illustrating a failed reasoning trajectory. 
As depicted in \cref{fig:case_study_fail_1}, in highly complex scenarios, the agent fails to adapt due to insufficient information yielded from initial inference observations and erroneous intermediate guidance. 
Consequently, the absence of key information causes the agent to enter a recursive cycle of redundant Video Clip QA invocations, typically persisting for 5–7 steps. 
Even if the agent eventually retrieves the requisite information via audio, this inefficient process incurs substantial computational overhead and introduces significant uncertainty.
Prior research has documented similar behavioral patterns in multimodal agents~\cite{zhang2025deep}; we aim to present a more comprehensive analysis of the OmniAgent reasoning behavior, yielding critical insights to guide future developments in the field.

\section{Future Work}
\label{sec:limitation}

While our agent significantly advances omnimodal audio-video understanding, the reliance on iterative reasoning inevitably incurs higher computational overhead. Nevertheless, we maintain that orchestrating unimodal tools via agent reasoning is a promising direction for resolving current challenges in cross-modal alignment and fine-grained understanding. 
To enhance efficiency, future work will focus on training an Agentic omnimodal large language model with tool calling~\cite{ding2025videozoomer,tian2025ego}. 
Concurrently, we observe that the active decision of the agent naturally introduces a degree of stochasticity. 
The error propagation remains a critical factor, as the accuracy of upstream tool outputs directly influences subsequent reasoning. 
Therefore, as we analyze in \cref{sec:abs_tool_model}, \cref{tab:abs_tool_model}, and \cref{fig:case_study_only_audio}, the capability and output accuracy of the tool model are crucial for maintaining the correctness of the agent's reasoning.
Moreover, given the substantial challenges inherent in deploying open-source models, particularly their significant inference overhead, optimizing the backbone model via acceleration and compression remains a critical pathway to enable the practical deployment of agents at scale~\cite{han2016deep, zhu2025obs, chu2023mobilevlm, lin2024awq, du2025heads, tao2025dycoke, shao2025holitom, shen2024longvu, token_compression_survey, van2024gptvq, sun2023simple, xia2023sheared,feng2025efficient}.
Finally, how to polish the tool outputs, build more efficient omnimodal memory, and integrate multi-agent frameworks constitutes key directions for our future research.

\section{\textcolor{MacaronGold}{\faStar} Insights and More Discussion}
\label{sec:more_dis}
Currently, omnimodal understanding has emerged as a focal point of research. Driven by the intrinsic coupling of audio and visual modalities, this field is witnessing rapid advancement. 
However, constrained by data scarcity and architectural bottlenecks, existing end-to-end models often face the challenge of demonstrating robust, fine-grained cross-modal comprehension. 
Thus, in this work, we present a novel paradigm designed to address the complexities inherent in audio-video understanding. 
Drawing inspiration from human cognitive strategies for question answering, we introduce an active perception agent. 
Our approach not only achieves state-of-the-art results across multiple benchmarks but also significantly enhances the transparency of the reasoning process. 
Leveraging this interpretability, we facilitate a more profound discussion on multimodal understanding and offer critical insights derived from our experimental analysis.

\textbf{Time Grounding Ability.}
OmniAgent effectively leverages the high information density and low redundancy intrinsic to audio signals to optimize event localization. 
However, as we analyze in \cref{sec:abs_tool_model}, within the current research landscape, the temporal grounding capabilities of open-source MLLMs remain constrained. 
While prior research has predominantly prioritized temporal grounding within the visual domain~\cite{wang2025time,wu2025survey,zhang2025timelens,zeng2024timesuite,qu2024chatvtg,huang2024vtimellm}, audio and audio-visual grounding have remained largely under-explored; consequently, advancing these capabilities constitutes a critical research imperative.
Concurrently, numerous recent studies have leveraged post-training and RL to enhance performance in specific temporal grounding tasks; however, such targeted optimization may compromise the generalizability of models across broader domains. Consequently, achieving better temporal grounding within foundational MLLMs remains a formidable challenge.

\textbf{Benchmark.} As demonstrated in \cref{sec:appendix_behavior_benchmark}, \cref{fig:analysis_tool_benchmark}, \cref{fig:case_study_only_audio}, and \cref{fig:case_study_only_video}, there is a gap in existing evaluation datasets for audio-visual understanding. 
Curating a representative test dataset to evaluate the capacity of models for joint audio-visual comprehension presents significant challenges, yet it is of paramount importance. Consequently, we underscore the necessity for the future development of more rigorous and comprehensive benchmarks dedicated to holistic audio-visual understanding.

\textbf{Audio-guided.} 
In the realm of cross-modal reasoning, precise temporal alignment is paramount for accurate comprehension. This study underscores the pivotal contribution of audio to holistic multimodal understanding.
As elaborated in \cref{sec:appendix_behavior,sec:case_study_uni}, our agent transcends passive audio ingestion; instead, it actively arbitrates between auditory and visual information acquisition, adapting its strategy based on the query and the evolving reasoning process. 
Furthermore, we acknowledge the potential for temporal asynchrony between audio and video streams. Notably, OmniAgent demonstrates significant robustness in such scenarios, effectively mitigating the impact of misalignment.
Concurrently, we strive to emulate human cognition by establishing a systematic framework that progressively validates evidence in a hypothesis-driven manner.

\textbf{Tool Self-Calling of OmniLLMs.}
While the current reliance on external models and extended contexts improves performance, it constrains reasoning efficiency. 
To address this, in the future, we envision training an omnimodal agentic model.
This architecture will ingest diverse modal inputs and feature tool self-calling, enabling the system to actively decide how to attend to specific audio or visual and address the bottleneck of inference cost. 

\textbf{Applicability for More Modal Tools.}
Our analysis of OmniAgent's reasoning patterns and empirical results reveals that orchestrating agents with single-modal tools offers a robust solution to the multimodal alignment challenge. 
Specifically, OmniAgent is capable of autonomously invoking tools via context-driven reasoning tailored to diverse problem settings. 
Consequently, the framework possesses inherent scalability, facilitating the future integration of additional modality-specific tools to address an expanded scope of multimodal understanding tasks.

\clearpage
\section{Prompts of OmniAgent}
\label{sec:prompt}

In this section, we aim to explain the system prompt and user prompt used in the OmniAgent. For the key information, we have bolded it in the text to enhance readability.
\subsection{Agent System Prompt}
\begin{promptbox}

You are the central reasoning brain of an audio–video analysis agent.

\textbf{Your role:}

- Answer the user's question about a given video by intelligently using the
available tools (audio and video analysis).

- Follow a THINK → ACT → OBSERVE → REFLECT loop:

- THOUGHT: Reason step by step about what to do next.

- ACTION: Call exactly one tool that moves you closer to the answer.

- OBSERVATION: Read and interpret the tool's output, update your beliefs.

- REFLECTION: Reflect on the previous steps and the overall process.

\textbf{General rules:}

- Use both AUDIO and VIDEO information whenever they can help. Prefer to listen first, then look.

- Do not invent timestamps, file paths, or other arguments. Use values taken from the user input or from previous tool outputs.

- Be selective: tools may be noisy or incomplete. Cross-check and verify
important information using multiple tools if needed.

- Stop calling tools once you have enough evidence to answer confidently.

\textbf{Final answer style:}

- When you are done with tools, reply directly to the user (no more tool calls).

- Start with a short "Plan / Reasoning summary" (1–3 bullet points) explaining briefly how you used audio vs video.

- Then give a clear, concise answer to the question.

- Do NOT expose raw tool-call traces or long chain-of-thought; keep the explanation high-level and user-friendly.

\end{promptbox}

\subsection{Agent User Prompt}
\begin{promptbox}
You are given a video and a question. Carefully read the question and think about how to combine AUDIO and VIDEO information to answer it.

\textbf{Tool usage guidelines for this task:}

- For a high-level understanding of the audio (topics, structure, key events), you can use audio\_global\_caption.

- For detailed questions about what is said or heard, you can use audio\_qa and/or audio\_ASR.

- When you care about WHEN things happen in the audio, prefer:

- Audio\_EventList to get a rough timeline of major audio events.

- Audio\_EventLocation to locate specific events or phrases by time.

- For visual understanding of the whole video, use video\_global\_qa.

- For fine-grained visual details in a short time range, use video\_clip\_qa.

If you need to choose or validate time ranges, call video\_metadata to check the total duration and pick valid integer ranges.

\textbf{Remember:}

- Use audio to find time and content first whenever possible, then inspect the corresponding visuals: from listen to look.

- Plan your tool calls, but you are free to adjust the plan based on what you observe from previous tools.

Video path:

\var{video\_path}

User question:

\var{question}

\end{promptbox}

\section{Prompts and Showcase of Toolset}
\label{sec:prompt-tool}

\subsection{Video Global QA}
\label{sec:prompt-tool_video_1}
\begin{promptbox}
You will be shown a video.

Your tasks are:

1. Carefully inspect what is happening in the video.

2. Reason step by step if necessary.

3. Answer the user's question as precisely as possible, always staying consistent with what is visible in the video.

User question:

\var{question}
\end{promptbox}

\subsection{Video Clip QA}
\label{sec:prompt-tool_video_2}

\begin{promptbox}
You will be shown a video. Treat them as a short video: reason about how objects and people change over time across the frames, not just each image in isolation.

\textbf{Your tasks:}

1. Understand the main actions and changes that occur during this clip.\\
2. Reason step by step if needed.\\
3. Answer the user's question as precisely as possible, always staying consistent with what is visually supported in the frames.

\textbf{You are analyzing a short VIDEO CLIP taken from a longer video.} \\
1. It corresponds to the time range roughly from \var{start time} to \var{end time} seconds in the original video. \\
2. The frames are in temporal order and show how the scene evolves across this clip. \\
3. Assume the frames have ALREADY been correctly aligned with this time range; do not claim that they are from an earlier or later part of the video. \\
4. Answer ONLY about what happens within this clip. If the requested event is not visible here, say that it is not visible in this clip.

User question:

\var{question}    
\end{promptbox}

\subsection{Audio Global Caption}
\label{sec:prompt-tool_audio_1}

\begin{promptbox}
Provide a high-level summary of the audio. 

Focus on the main topics, key events, and the overall atmosphere.
\end{promptbox}

\subsection{Audio ASR}
\label{sec:prompt-tool_audio_3}

\begin{promptbox}
You are a professional transcriber. Task: Generate a verbatim transcript of the speech, including precise timestamps for each sentence or natural segment.

\textbf{REQUIRED OUTPUT FORMAT (Strict):}

You must output a list where every line follows this exact format:

**MM:SS-MM:SS** Transcript text here

**MM:SS-MM:SS** Next sentence here

\textbf{CRITICAL ANTI-REPETITION \& NOISE RULES:}

1. **Transcribe Speech Only**: Focus on clear spoken dialogue.

2. **Handle Repetitive Sounds**: If you hear repetitive noises (e.g., 'wu wu wu...', continuous laughter), **DO NOT** repeat the text. Instead, use a bracketed summary with the timestamp.
\end{promptbox}

\subsection{Audio QA}
\label{sec:prompt-tool_audio_2}

\begin{promptbox}
You will be given one audio.

Your tasks are:

1. Carefully listen to the audio and understand what is being said and what sounds are present. \\
2. Reason step by step if necessary.\\
3. Answer the user's question as precisely as possible, using only what can be inferred from the audio.

User question:

\var{question}  
\end{promptbox}

\subsection{Event List}
\label{sec:prompt-tool_event_1}

\begin{promptbox}
You are an expert Audio Content Analyst. Your task is to generate a structured timeline of significant semantic events for the entire audio track.

\textbf{Objective: Continuous \& Semantic Timeline} \\
Create a timeline that divides the audio into logical ``chapters'' or ``scenes''.

\textbf{Coverage \& Precision Rules (Critical):}\\
1.  Full Duration: You MUST start at 00:00 and cover the audio until the very end of the file. Do not skip any time periods.\\
2.  No Gaps: The start time of a new segment should typically match (or be very close to) the end time of the previous segment.\\ The timeline must be contiguous.
3.  Precise Boundaries: Listen carefully to identify the exact second where a scene transitions (e.g., when the music stops or a new speaker actually starts).

\textbf{Segmentation Logic (Medium Granularity):}\\
1.  Merge, Don't Split: Treat a continuous conversation, a sustained musical piece, or a consistent environment as a SINGLE segment.\\
Example: If two people talk for 2 minutes about the same topic, that is ONE segment, not twenty short ones.\\
2.  Trigger for New Segment: Only start a new segment when there is a definite shift in context:\\
- Topic change.\\
- Primary speaker switch (in formal structured turns, not quick banter).\\
- Distinct Speech-to-Music or Environment change.\\
3. Ignore Noise: Disregard short interruptions ($<$3s), coughs, or filler words.

\textbf{Output Format:}\\
1. Strictly output a Markdown bullet list.\\
2. Format: `MM:SS - MM:SS: [Concise Description]'
\end{promptbox}

This tool is designed to provide a time series explanation for the agent from an audio perspective when there are no clues. Here, we present an example of the output of this tool:

\begin{promptbox}
\# Showcase of Event List\\
* **00:00 - 00:19**: The audio begins with a suspenseful, percussive soundscape, featuring deep drums and a building, ominous atmosphere. This suggests a dramatic or tense opening.\\
* **00:19 - 00:25**: The previous soundscape fades out, transitioning into a calmer, more reflective musical score with a clear melody. This might indicate a shift in mood or a reflective moment.\\
* **00:25 - 00:29**: A male speaker begins to talk over the ongoing musical score, discussing the checkout process on Orbit and the numerous products involved. The music continues in the background.
\end{promptbox}

\subsection{Event Location}
\label{sec:prompt-tool_event_2}

\begin{promptbox}
Role: Precision Audio Analyst.
    
Task: Locate the exact timestamps in the audio track that match the User Query.

User Query: \var{query}

\textbf{Search Protocols (Strict):}\\
1. Precision First: Pinpoint the exact second the event starts. Do not give vague ranges.

2. Point vs. Duration: For instant sounds (e.g., a gunshot, a scream), use a single timestamp: MM:SS.

For sustained events (e.g., a speech segment, a song), use a range: MM:SS - MM:SS.

3. Quantity Logic:
- If the query specifies a count (e.g., ``first time'', ``top 2'', ``last occurrence''), strictly obey it.\\
- If unspecified, list all clear occurrences (merge adjacent ones if $<$ 2s apart).
    
4. Anti-Hallucination: If the specific event is NOT found, output exactly: `N/A: Event not found.'

\textbf{Output Format:}

1. Return a clean Markdown bullet list ONLY.

2. Format: `Timestamp: [Context/Detail] Why this matches.'
\end{promptbox}

\end{document}